\documentclass[utf8]{frontiersSCNS} 

\usepackage{url,hyperref,microtype,subcaption}
\usepackage[onehalfspacing]{setspace}
\usepackage{listings}
\usepackage{csquotes}

\usepackage{todonotes}

\def\keyFont{\fontsize{8}{11}\helveticabold }
\def\firstAuthorLast{Sample {et~al.}} 
\def\Authors{Karine Miras\,$^{1,*}$, Eliseo Ferrante$^{1}$, A.E. Eiben\ $^{1}$ }
\def\Address{$^{1}$Computer Science Department\\ Vrije Universiteit Amsterdam, Amsterdam, Netherlands}
\def\corrAuthor{Corresponding Author}

\def\corrEmail{k.dasilvamirasdearaujo@vu.nl }

\begin{document}
\onecolumn
\firstpage{1}

\title[Environmental regulation using \textit{Plasticoding} for the evolution of robots]{Environmental regulation using \textit{Plasticoding} for the evolution of robots} 

\author[\firstAuthorLast ]{\Authors} 
\address{} 
\correspondance{} 

\extraAuth{}

\maketitle

\begin{abstract}

Evolutionary robot systems are usually affected by the properties of the environment indirectly through selection. In this paper, we present and investigate a system where the environment also has a direct effect: through regulation. We propose a novel robot encoding method where a genotype encodes multiple possible phenotypes, and the incarnation of a robot depends on the environmental conditions taking place in a determined moment of its life. This means that the morphology, controller, and behavior of a robot can change according to the environment. Importantly, this process of development can happen at any moment of a robot lifetime, according to its experienced environmental stimuli. We provide an empirical proof-of-concept, and the analysis of the experimental results shows that \textit{Plasticoding} improves adaptation (task performance) while leading to different evolved morphologies, controllers, and behaviour.    
\tiny
 \keyFont{ \section{Keywords:} evolutionary robotics, morphological evolution, phenotypic plasticity, environmental regulation, locomotion, environmental effects} 
\end{abstract}

\section{Introduction}


What makes natural life remarkably complex goes beyond having genes encoding a trait or behavior, as it concerns also mechanisms in the DNA that regulate the expression of these genes as a function of environmental conditions. That is, genes should be activated `at the right place at the right time'. An amazing number of 95\% of DNA does not code for any protein: Part of it is responsible for regulation~\footnote{Another part of the DNA is often denoted as ``junk'', that is, it has either no function or (mostly likely) it has a function that we do not know yet.}~\citep{sapolsky2017behave}. In fact, the more genomically complex an organism, the larger the percentage of its DNA devoted to environmental regulation~\citep{sapolsky2017behave}. This regulation happens through a process once called \textit{epigenetics}~\citep{bossdorf2008epigenetics}, a term that recently has been utilized only in cases when this regulation results in heritable regulatory changes~\citep{sapolsky2017behave}. One of the results of this regulation is lifetime \textit{phenotypic plasticity}, and it concerns the capacity of an individual to develop aspects of its phenotype, such as morphology, physiology, synaptic connections, in response to given environmental stimuli during its lifetime~\citep{fusco2010phenotypic}. 

Although phenotypic changes like learning~\citep{fusco2010phenotypic} and training~\citep{kelly2011phenotypic} are also examples of \textit{phenotypic plasticity}, here we consider only phenotypic changes that happen through regulation. \textit{Phenotypic plasticity} is pervasive in nature and may accelerate, decelerate, or have an insignificant effect on evolutionary change~\citep{price2003role}. Some examples of lifetime \textit{phenotypic plasticity} acting on the body are Passerine birds that change their musculature to cope with winter~\citep{liknes2011phenotypic}, and several vertebrate species that suffer color changes in different seasons~\citep{mills2018winter}. As for behavioral changes caused by environmental regulation acting on the brain, think of the physiology of a mother changing to produce milk when she smells her baby. This is a change that happens through the activation of genes. More specifically, this example is thoroughly described by~\citet{sapolsky2017behave} as ``A female smells her newborn, meaning that odorant molecules that floated off the baby bind to receptors in her nose. The receptors activate and (many steps later in the hypothalamus) a transcription factor activates, leading to the production of more oxytocin. Once secreted, the oxytocin causes milk letdown. Genes are not the deterministic holy grail if they can be regulated by the smell of a baby’s tushy. Genes are regulated by all the incarnations of environment. Promoters and transcription factor introduce \textit{if/then clauses}: \textit{If} you smell your baby, \textit{then} activate the oxytocin gene.''

Within engineering, the research areas related to artificial evolution are that of Evolutionary Computing \citep{eiben2003introduction, eiben2015evolutionary} and Evolutionary Robotics ~\citep{nolfi2000evolutionary, nolfi2016evolutionary, doncieux2015evolutionary}. These fields have addressed the evolution of robot controllers (brains) with considerable success but evolving the morphologies (bodies) has received much less attention. Importantly, the influence of the environment has been even more scarcely investigated. Moreover, although it is not uncommon to use developmental robot DNA structures, there is no substantial work in the literature that successfully allows a genotypic structure to be regulated by changes in environmental conditions.

The key idea of this paper is to develop a novel robot DNA structure, that is, a new encoding method that endows robots with lifetime \textit{phenotypic plasticity}. We achieve this through a genotype-phenotype mapping that responds and is modified according to the environmental conditions at given moments of a robot life. This idea represents a significant departure from existing systems, where the genotype-phenotype mapping is \enquote{injective}, that is, each genotype encodes only one possible phenotype. This holds for both direct and indirect (e.g. generative or developmental) mappings~\citep{rothlauf2006representations}.  In contrast, here we study genotype-phenotype mappings where a genotype encodes multiple possible phenotypes and the actual `incarnation' (the robot body and brain) depends on the environment. 

The expected benefits of plasticity include higher efficiency and efficacy of robot evolution, together with increased responsiveness to environmental changes. We expect increased efficiency (speed) because an informed genotype-phenotype mapping makes reproduction less blind. Hence, the total number of trials (new robots born over the course of evolution) to evolve good robots should be lower than in systems using the conventional representations. Efficacy is the other side of the same coin: given a fixed search budget (maximum number of trials for evolution), an informed genotype-phenotype mapping will expectedly achieve better solutions. Last, but not least, a robot population that is equipped with an environment dependent genotype-phenotype mapping can cope with environmental changes better than a system where adaptation is induced through selection only. 
The specific objectives of this paper are: 
\begin{enumerate}

  \item  To design a novel robot encoding with the capacity of \textit{phenotypic plasticity} during the robot lifetime. We call this robot encoding \textit{Plasticoding}. 
  \item Use this robot encoding to achieve better adaptation (i.e., improvement in task performance) to seasonal environmental conditions than a baseline (non-plastic) robot encoding.
 
\end{enumerate}

Additionally, we investigate this improvement in performance through answering the following research questions:
\begin{enumerate}

\item  What is the effect of \textit{phenotypic plasticity} on the \textit{morphological} properties?
\item  What is the effect of \textit{phenotypic plasticity} on the \textit{controller} properties?
\item  What is the effect of \textit{phenotypic plasticity} on the emergent \textit{behavior}?
\end{enumerate}

\section{Related Work}
Existing work related to the evolution of virtual creatures dates back to the 1990s, when morphological (additionally to controller) evolution was addressed by Sims~\citep{sims1994evolving}. Sim's work was later put on a more solid footing by Pfeifer and Bongard~\citep{pfeifer2005morphological}.

In~\citep{bongard2011morphological} it has been shown that ontogenetic, i.e., lifetime development, can not only accelerate the discovery of successful behavior, but also produce robots that are more robust to variations of environmental conditions.
~\citet{2014-AB} utilized an information-theoretic measure of complexity to assess virtual creatures evolved in a vast range of environments. The authors demonstrated that increasing the complexity of the environmental conditions might result in an increase to the morphological complexity of the creatures.

A developmental mechanism being presented as epigenetics was proposed in~\citep{brawer2017epigenetic}, but in fact it was not dependent on environmental influences. 
The effect of different developmental mechanisms was studied in~\citep{kriegman2018interoceptive} by changing the stiffness of soft robots according to environmental changes. However, no improvement to evolvality was achieved through it. A similar investigation was presented in~\citep{kriegman2018morphological}, this time obtaining improvements in evolvability. Nevertheless, although both these studies concern lifetime development mechanisms, the regulatory environmental changes were caused by the displacement of the robot itself, and therefore no actual `changing' environmental conditions were considered while robots evolved always in a flat plane. In~\citep{daudelin2018integrated}
reconfigurable robots were evolved to cope with actual changes in the environmental conditions as they moved about, but no quantification of this effect on the morphological level was provided.

\section{Methods}
In our methodology, we use modular robots to represent the morphology (see Section~\ref{sec:morphology}) and neural networks to represent the controllers (Section~\ref{sec:controllers}). These two together represent the phenotypes, as they express the traits that ultimately, through the interaction with the environment, determine fitness. The evolutionary process acts on a higher level, the level of the genotypes, whose representation is explained in Section~\ref{sec:representation}. 

\begin{figure}[t]
\begin{center}
\includegraphics[width=5.5in]{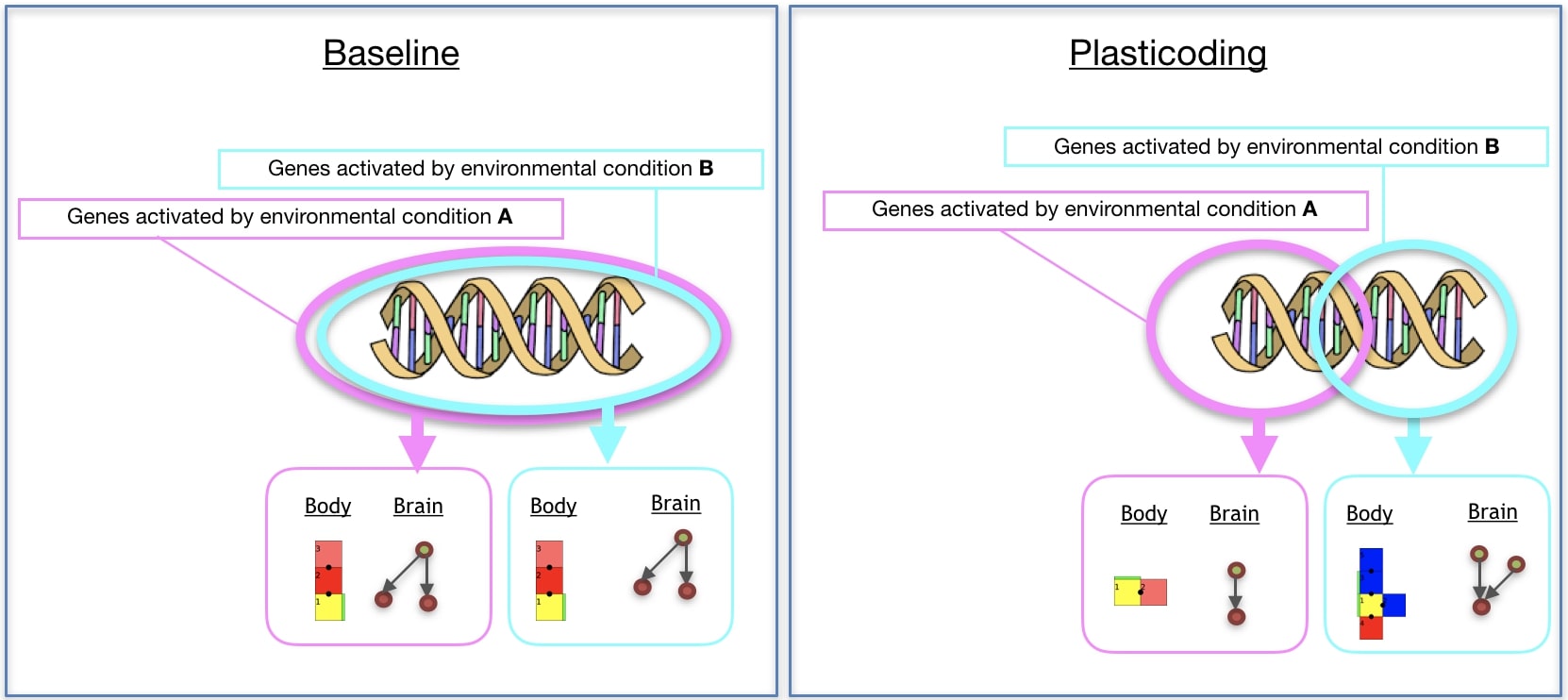}
\caption{Effects of environmental regulation on phenotype.}
\label{fig:regulation3}
\end{center}
\end{figure}

\begin{figure}[t]
\begin{center}
\includegraphics[width=4in]{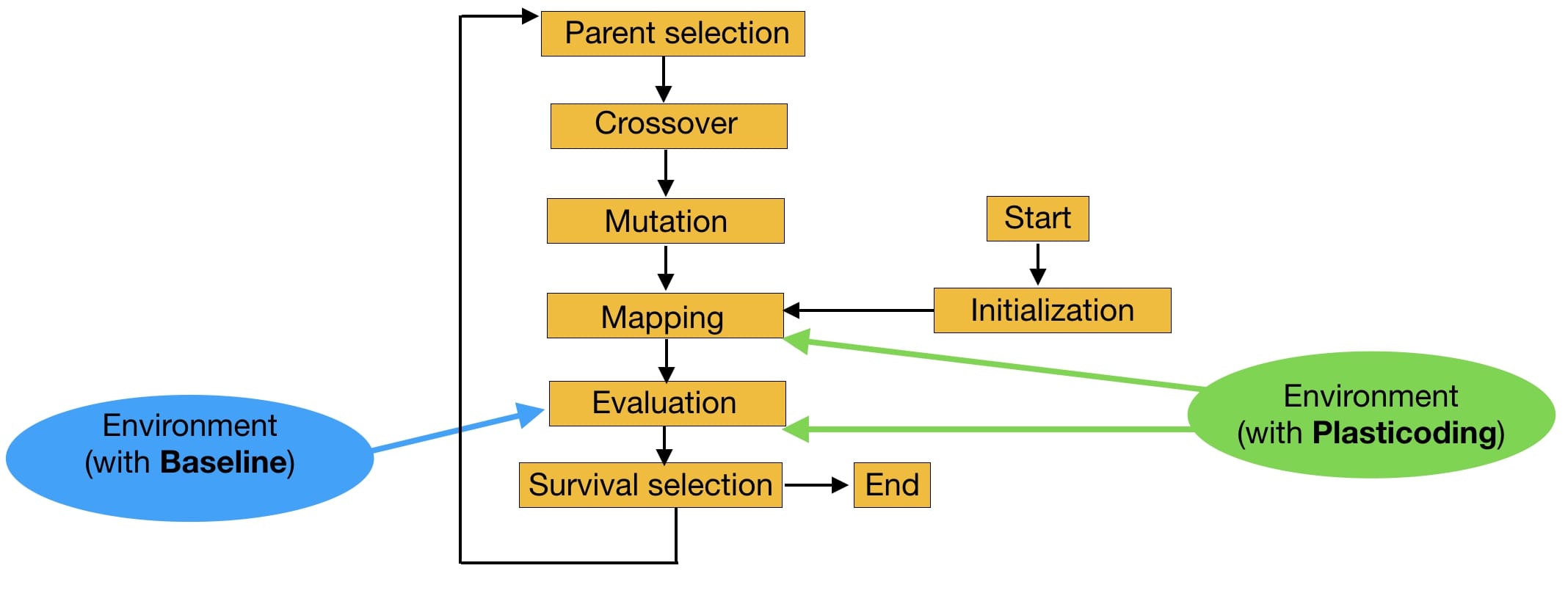}
\caption{Effect of the environment on the evolutionary process of the population.}
\label{fig:regulation2}
\end{center}
\end{figure}

In this paper, we extend a robot encoding we proposed in previous work~\citep{miras2019effects}. Here, we refer to the previous encoding as \textit{Baseline}, and to the new encoding as \textit{Plasticoding}. The differential added by the \textit{Plasticoding} concerns environmental regulation, allowing an individual to develop a different phenotype, i.e., morphology and/or controller, according to the conditions of the environment it is in (Fig.~\ref{fig:regulation3}). While for the \textit{Baseline} the environment acts on the stage of the evaluation of the robots, for the \textit{Plasticoding} it acts also on the stage of mapping the genotype to the phenotype (Fig.~\ref{fig:regulation2}). The methodology used for the regulatory mechanism is explained in Section~\ref{sec:meth_regulation}.

Genotypes are converted into phenotypes through a mapping process, which is explained in Section~\ref{sec:genopheno}. In the first generation, the genotype of the initial population is initialized according to the procedure described in Section~\ref{sec:genoinit}.  During the evolutionary process, the operators of crossover and mutation are applied, which are explained respectively in Section~\ref{sec:crossover} and Section~\ref{sec:mutation}. The overall evolutionary process is explained in Section~\ref{sec:evolution}.

\subsection{Morphology}
\label{sec:morphology}
\begin{figure}[t]
\begin{center}
\includegraphics[width=2.5in]{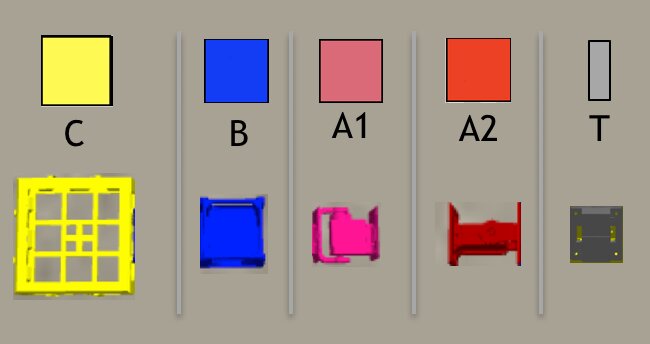}
\includegraphics[width=1.75in]{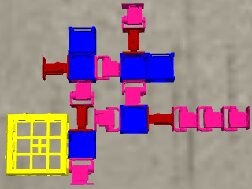} 
\caption{On the left, the robot modules: Core-component with controller board (C); Structural brick (B); Active hinges with servo motor joints in the vertical (A1) and horizontal (A2) axes; and Touch sensor (T).  Modules C and B have attachment slots on their four lateral faces, and A1 and A2 have slots on their two opposite lateral faces; T has a single slot which can be attached to any slot of C or B.  
On the right, an example of simulated robot.}
\label{fig:components}
\end{center}
\end{figure}

Each  morphology phenotype (a `body') is composed of modules~\citep{auerbach2014robogen} as shown in Fig~\ref{fig:components}. Each module has a cuboid shape, and has slots where other modules can attach. The morphologies can only develop in 2 dimensions, that is, the modules do not allow attachment to the top or bottom slots, but only to the lateral ones. There are five different types of modules, as reported in Table~\ref{table1}: core components, bricks, vertical joints, horizontal joints, and touch sensors. Any module can be attached to any module through its slots, except for the touch sensors, which cannot be attached to joints. Each module type is represented by a distinct symbol (see Table~\ref{table1}), and this is also the same language used in the genotype representation, described in Section~\ref{sec:representation}.  
 
\vspace{0.1cm}
\subsection{Controller}
 \label{sec:controllers}
A controller phenotype (a `brain') is a hybrid artificial neural network (Fig~\ref{fig:decoding}, right), which we call Recurrent Central Pattern Generator Perceptron ~\citep{miras2019effects}. This network is formed by two types of nodes, i.e., input nodes associated with the sensor modules, and oscillator neuron nodes associated with the joint modules. For every joint in the morphology, there exists a corresponding oscillator neuron in the network, whose activation function is defined by Eq.~\eqref{eq:sine}, which represents a sine wave defined by amplitude, period, and phase offset parameters. This activation function adjusts the output to fit the range of our servo motors, as proposed in~\citep{hupkes2018revolve}.

\begin{equation}
\label{eq:sine}
      O = 0.5 - \frac{a}{2} +  \frac{\sin\Bigg(\frac{2 * \pi}{p} * (t - p * o))\Bigg) + 1}{2}  * a 
\end{equation}

\noindent where, $t$ is the time step, $a$ is the amplitude, $p$ is the period, and $o$ is the phase offset. The parameters $a$, $p$, and $o$ can vary from $0$ to $10$.
The different oscillator neurons are not directly interconnected, and every oscillator neuron may or may not possess a direct recurrent connection, according to the rules that evolve. 

Additionally, for every sensor in the morphology, there exists a corresponding input in the network, and each input might connect to one or more oscillator neurons. The oscillator neurons generate a constant pattern of movement, even if the robot is not sensing anything, so that the sensor inputs can be used either to reduce or to reinforce movements. 

 \begin{table}[t]
\caption{  
{\bf Alphabet of the grammars. Terminology is explained in Section~\ref{ref:maplatedev}.}}
\begin{tabular}{|l|l|}
\hline
\multicolumn{2}{l}{\bf Modules} \\ \hline
 
\textbf{C}  &  core-component  (axiom $w$)\\ \hline
\textbf{B}  &  brick  \\ \hline
\textbf{A1($w_{v}, a_v, p_v, o_v$)}  &  vertical joint  \\ \hline
\textbf{A2($w_{h}, a_h, p_h, o_h$)}  &  horizontal joint  \\ \hline
\textbf{T($w_t$)}  &  touch sensor  \\ \hline
 
\multicolumn{2}{l}{ $w_{v},w_{h}, w_t $   are sampled from a uniform distribution ranging from $-1$ to $1$}\\
\multicolumn{2}{l}{ $ a_v, p_v, o_v, a_h, p_h, o_h $ are sampled from a uniform distribution ranging from $1$ to $10$}\\

  \hline
\multicolumn{2}{l}{\textbf{Morphology-mounting commands}} \\ \hline 
\textbf{add\_right}  & add new module to the right of  \textit{module-reference}\\ \hline
\textbf{add\_front}  & add new module to the front \textit{module-reference}  \\ \hline
\textbf{add\_left}  & add new module to the left of \textit{module-reference}  \\ \hline

\multicolumn{2}{l}{\textbf{Morphology-moving commands}} \\ \hline 
\textbf{move\_back}  & move \textit{module-reference} to the module at the back of \textit{module-reference}  \\ \hline
\textbf{move\_right}  & move \textit{module-reference} to the module at the right of \textit{module-reference}     \\ \hline
\textbf{move\_front}  & move \textit{module-reference} to the module at the front of \textit{module-reference}    \\ \hline
\textbf{move\_left}  & move \textit{module-reference} to the module at the left of \textit{module-reference}   \\

\hline
\multicolumn{2}{l}{\textbf{Controller-moving commands}} \\ \hline 
\textbf{move\_ref\_I($t_i, d_i$)}   &  update \textit{input-reference} with the input connected to edge $d_i$ of \\    \multicolumn{2}{c}{ the neuron connected to edge $t_i$ of \textit{input-reference} }  \\ \hline
\textbf{move\_ref\_N($t_n, d_n$)}   &  update \textit{neuron-reference} with the neuron connected to edge $d_n$ of \\    \multicolumn{2}{c}{ the input connected to edge $t_n$ of \textit{neuron-reference} }  \\ \hline
 
   \multicolumn{2}{l}{
  $t_i = \lceil\sqrt{v_1^2)}\rceil $ and $t_n = \lceil\sqrt{v_2^2)}\rceil $, and they are used to move the reference to a temporary node} \\ 
 \multicolumn{2}{l}{ $d_i = \lceil\sqrt{v_3^2)}\rceil $ and $d_n = \lceil\sqrt{v_4^2)}\rceil $, and they are used to move the reference to a definite node} \\  
    \multicolumn{2}{l}{ $v_1, v_2, v_3, v_4$ are sampled from a normal distribution with $\mu=0$ and $\sigma=1$ } \\
 
  \multicolumn{2}{l}{ If any of $t_i$, $d_i, t_n, d_n$ is greater than the number of edges of its corresponding node,} \\
   \multicolumn{2}{l}{its value is updated with this number of edges.  }  \\ \hline
 
\multicolumn{2}{l}{\textbf{Controller-changing commands}} \\ \hline 
\textbf{add\_edge($w_{e1}$)}   &  add an edge  between \textit{input-reference} and \textit{neuron-reference}  \\ \hline
\textbf{loop($w_l$)}   &  add a recurrent edge to \textit{neuron-reference} \\ \hline
   \multicolumn{2}{l}{ $w_{e1}, w_l$   are sampled from a uniform distribution ranging from $-1$ to $1$ } \\ \hline
   
\textbf{mutate\_edge($w_{e2}$)}   &  mutate the weight of the edge between \textit{input-reference}  and \textit{neuron-reference} \\ \hline
\textbf{mutate\_amp($m_a$)}   &  mutate amplitude of \textit{neuron-reference} \\ \hline
\textbf{mutate\_per($m_p$)}   &  mutate period of \textit{neuron-reference} \\\hline
\textbf{mutate\_off($m_o$)}   &  mutate phase offset of \textit{neuron-reference} \\ 
\hline
   \multicolumn{2}{l}{
    $w_{e2}, m_a, m_p, m_o$   are sampled from a normal distribution with $\mu=0$ and $\sigma=1$  } \\  \hline
 
\end{tabular}

\label{tab:alphabet}
\label{table1}
 
\end{table}

\subsection{Genotype Representation}
\label{sec:representation}

Our robot genotype is a generative model, and is represented with an L-System inspired in~\citep{hornby2001body}, conjointly encoding both morphology and controller. L-Systems are parallel rewriting systems~\citep{jacob1994genetic} composed by a grammar defined as a tuple $G = (V, w, P)$, where
 
\begin{itemize}
  \item $V$, the alphabet, is a set of symbols containing replaceable and non-replaceable symbols.
  \item $w$, the axiom, is a symbol from which the generative process starts.
  \item $R$ is a set of regulatory tuples ($c$, $p$) for the replaceable symbols, where $c$ is a regulation clause and $p$ is a production-rule. 
  
\end{itemize}
 
Each genotype is a distinct grammar, making use of the same alphabet (Table~\ref{tab:alphabet}), and the alphabet is formed by symbols that represent types of morphological modules as well as commands for assembling modules together and others for defining the structure of the controller. The symbols in the category Modules are replaceable, while the symbols of all other categories are non-replaceable.

\subsection{Regulation clauses}
\label{sec:meth_regulation}

\citet{sapolsky2017behave} coins a metaphor for the environmental regulation, calling regulation factors as ``\textit{if then clauses}''. Here, we abstract this metaphor, and implement it in a literal sense. This way, for us a regulation clause is a Boolean expression, which is denoted by $c$ in the tuple ($c$, $p$). In \textit{Plasticoding}, each clause contains up to $u$ terms (in our case $u=2$), and one same term may be repeated in the clause. Each term represents a comparison between an environmental state that can be sensed by the robots and a value that is \textit{True} or \textit{False}. The $u$ terms are combined using \textit{and} and \textit{or} operators.  Additionally, in \textit{Plasticoding} every replaceable symbol from $V$ appears in exactly $l$ tuples (in our experiments we limited the study $l=2$), and the selection of the production-rules to be used during development depends on the activation resulting from regulation clauses. In contrast, \textit{Baseline} is a special case: because there is no regulation, the clause $c$ is always \textit{True} and consequentially every replaceable symbol in $V$ can only appear in one of the tuples and therefore has only one production-rule associated to it.

The environment is the element that determines which production rules are activated through the regulation clauses. Although  in the current experiments we utilize only one environmental state, that describes the inclination of the ground in the environment (represented by a term called \emph{inclined}), our methodology could be used to describe multiple environmental states.  The \emph{inclined} environmental state can be sensed by the inertial measurement unit (IMU) sensor of the robots, which provides data about the orientation of the robot in space. If the robot's center of mass has zero inclination, the term \emph{inclined} assumes the value \textit{False}, otherwise it assumes the value \textit{True}. A few didactic examples of regulation clauses are listed below:

\noindent \textbf{Ex.1:} \textit{if $inclined$ = True then ...} \\
\textbf{Ex.2:}  \textit{if $inclined$ = True or $inclined$ = False then ...} \\
\textbf{Ex.3:}  \textit{if $inclined$ = False and $hot$ = True then ... }\footnote{This example shows how future extensions with multiple environmental states may look like.}\\

\subsection{Genotype-phenotype mapping}
 \label{sec:genopheno}
For the \textit{Baseline} method, the mapping from genotype to phenotype (the development), plays out in two stages that we call, respectively, \emph{early} and \emph{late} development. 
For the \textit{Plasticoding}, the mapping plays out in three stages, with the regulation stage preceding the \emph{early}  development stage.

\subsubsection{Environmental regulation}

\begin{figure}[t]
\begin{center}
\includegraphics[scale=0.3]{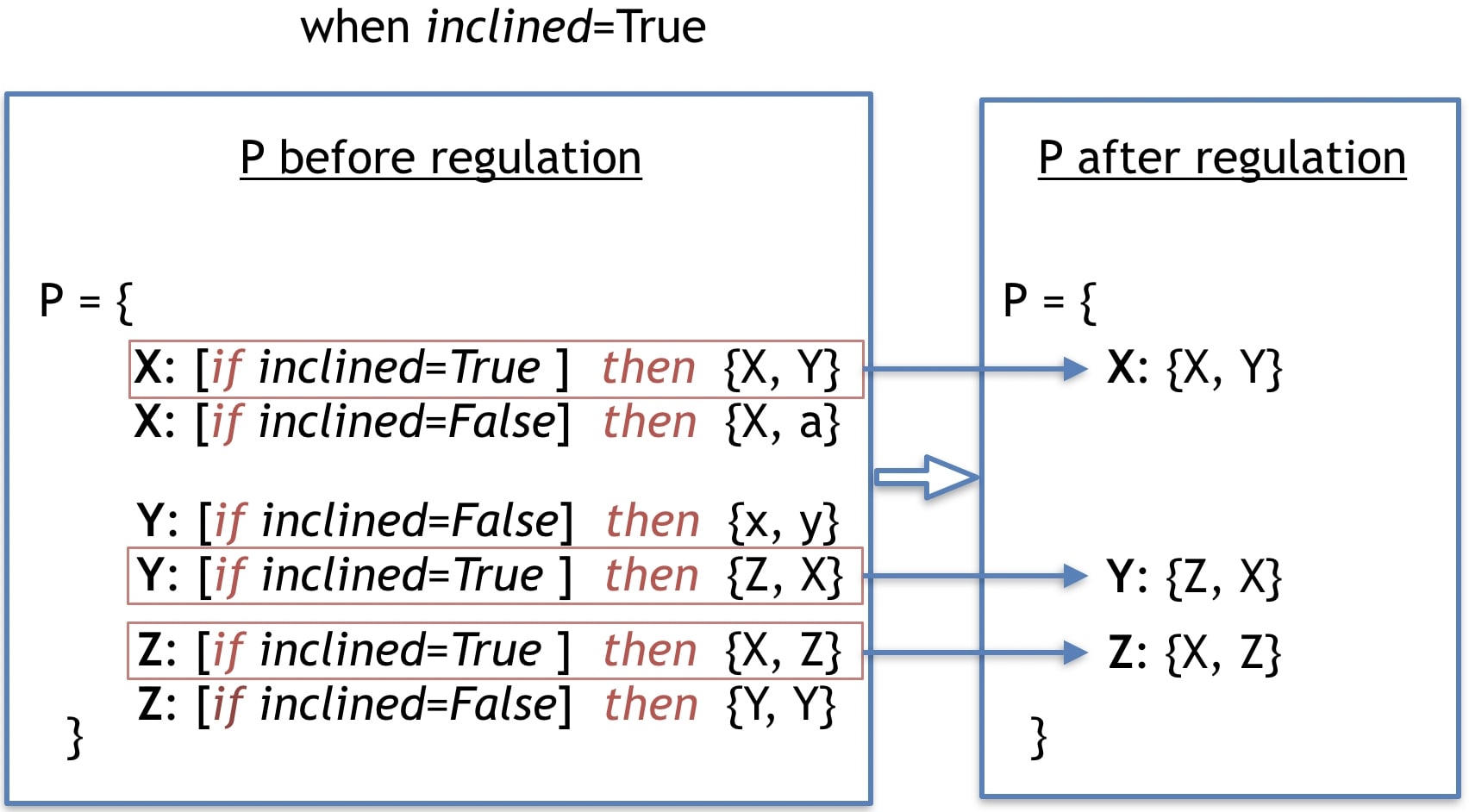}
\caption{This example shows the process of regulation of a genotype, selecting its rules to be activated due the environmental state. The term used to define the environmental state is called \textit{inclided}, and was sensed to be \textit{True} when the regulation process took place. The production-rules in the second frame are the ones that became true given the environmental state, and thus were selected go be in the final set $P$, which will be utilized in the \textit{early} development.}
\label{fig:regulation}
\end{center}
\end{figure}

The environmental regulation stage is responsible for selecting the production-rules that should be active during the \textit{early} development stage, according to the environmental states sensed by the robot. In the case of the \textit{Baseline}, all production-rules are always active, because there is no regulation. Therefore, effectively, this stage does not occur at all in \textit{Baseline}. In the case of the \textit{Plasticoding}, for a production-rule to be active, its regulation clause must be \textit{True}. Because multiple regulation clauses for one replaceable symbol can be \textit{True}, it is possible that multiple production-rules are activated. In this case, the multiple production-rules are concatenated sequentially as a single production-rule. Conversely, once multiple regulation clauses can be \textit{False}, it is possible that no production-rule gets activated. In this case the replaceable symbol is not replaced. Figure~\ref{fig:regulation} depicts an example of a process of regulation of a genotype.

\subsubsection{Mapping stage 1: early-development}
 
The axiom $w$ of the grammar is rewritten into a more complex string of symbols according to the activated production-rules of the grammar. During the rewriting, for a number of iterations $k=3$, each replaceable symbol is simultaneously replaced by the symbols of its active production-rules. The following didactic example depicts the process of rewriting of our L-System representing one possible genotype, i.e., grammar. For the case of the \textit{Plasticoding} method, it is assumed in this example that the regulation which activates production-rules has already been carried out.

    \begin{lstlisting}
          w = X 
          V = {X, Y, Z, a}
          P = {  
                X: {X,Y},  
                Y: {Z, a},  
                Z: {X, Z}  
            }
    \end{lstlisting}
  
Given the above grammar, the rewriting follows as: 
  
\begin{lstlisting}
          Iteration 0: X 
          Iteration 1: X Y 
          Iteration 2: X Y Z a 
          Iteration 3: X Y Z a X Z a
\end{lstlisting}
 
The final string will contain non-replaceable symbols (Modules) and replaceable symbols (everything else). All these symbols can be interpreted with the process described hereafter.
 
\subsubsection{Mapping stage 2: late-development}
\label{ref:maplatedev}
 
The early-developed phenotype from stage 1 is an intermediate phenotype made as a string of symbols, which must be mapped (late-developed) into a final phenotype. To aid the process of construction of the late-developed phenotype, multiple positional references (turtles) are kept: a) a reference to the current module in the morphology, that we call a \textit{module-reference}; b) a reference to the current oscillator neuron of the neural network of the controller, that we call a \textit{neuron-reference}; c) a reference to the current sensor input of the neural network of the controller, that we call an \textit{input-reference}; a reference to which slot of the current module a new module should be attached to, that we call a \textit{slot-reference}.

 \begin{figure}[t]
\begin{center}
\includegraphics[width=5.1in]{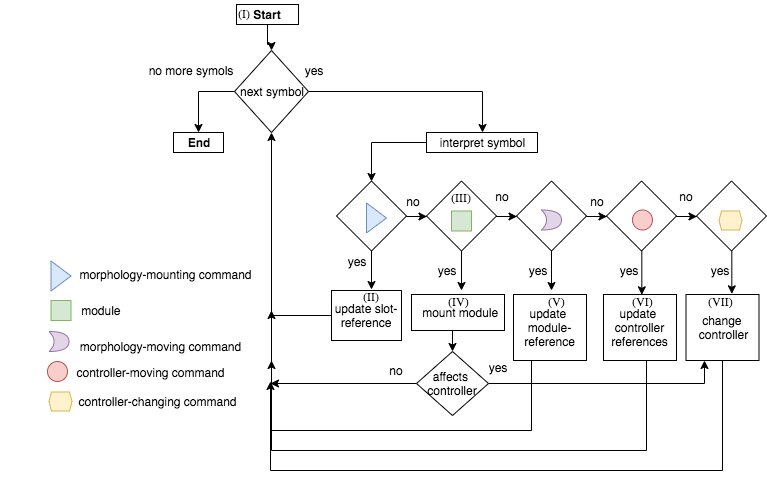}
\caption{Flow-chart of the late-development process. From the left to right of the string, each symbol of the early-developed phenotype (string) goes thorough this process, being interpreted and developed (or not expressed).}
\label{fig:latedev}
\end{center}
\end{figure}

\begin{figure}[t]
    \begin{center}
      \includegraphics[width=5.3in]{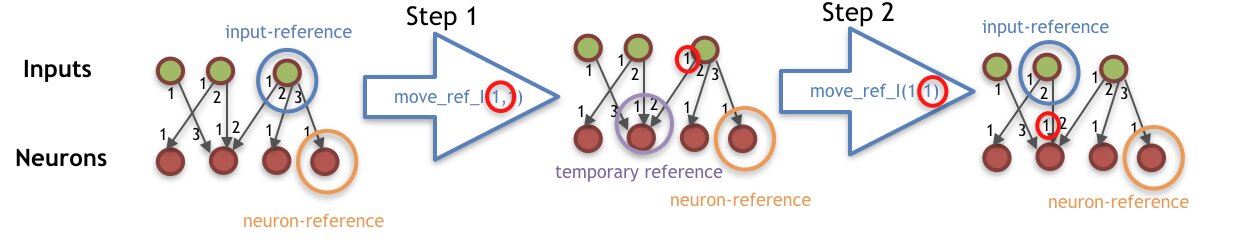}
    \caption{Illustration of command \textbf{move\_ref\_I($t_i, d_i$)}, having $t_i=1$ and $d_i=1$. The procedure of the command \textbf{move\_ref\_N($t_n, d_n$)} is analogous to this.}
        \label{fig:fts}
    \end{center}
    \end{figure}

\begin{figure}[t]
\begin{center}
\includegraphics[width=5in]{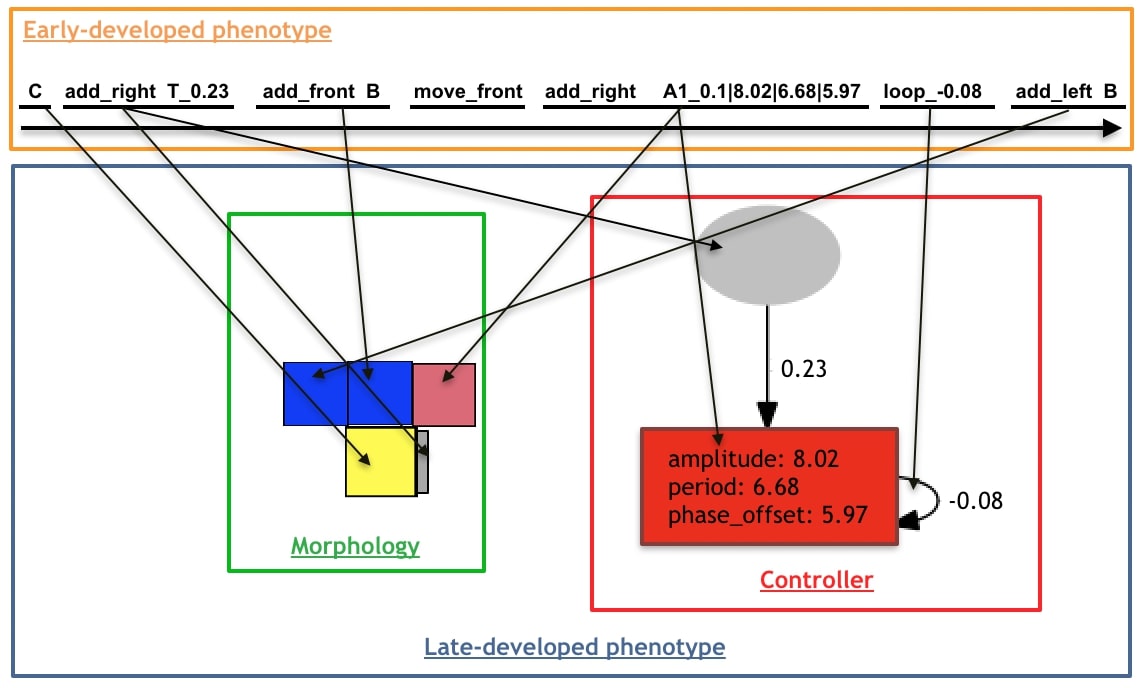}
\caption{Process of decoding an early-developed phenotype into a late-developed phenotype with morphology and controller. From the left to right of the string, symbols are interpreted and developed, making incremental changes to the phenotype. An arrow going from the genotype to the phenotype should be interpreted as the process leading to the creation of the phenotype component pointed at by the arrow after the interpretation of the genotype component at the starting end of the arrow.}
\label{fig:decoding}
\end{center}
\end{figure}

From the beginning until the end of the string, each symbol is interpreted and developed. Nonetheless, for multiple reasons explained below, it is possible that a symbol ends up not being expressed in the phenotype. Furthermore, a maximum amount of $m$ modules is allowed in a morphology, so that during late-development, after reaching this maximum, any upcoming modules are not expressed in the phenotype. The late-development of the phenotype for morphology and controller is depicted in the flowchart of Fig~\ref{fig:latedev}, and detailed hereafter, where we reference parts of this flowchart through Roman numerals:

\begin{itemize}

 \item \textbf{$I$}: Because the first symbol of the string is always $C$, it is the first module to be added to the morphology, and the \textit{module-reference} is updated with it. At this moment, the references of left, front, right, and back of the turtle are, respectively, left, up, right, and down (for a robot seen from top-down). 

 \item \textbf{$II$}: The interpretation of any Morphology-mounting command updates the \textit{slot-reference} with the slot indicated by the command. If the \textit{slot-reference} is not empty, it is overwritten, meaning that the command used for setting this previous slot into the reference is not expressed.  
 
 \item \textbf{$III$}: If the symbol is a module, it is coupled with the command in the \textit{slot-reference} (if there is one).

 \item \textbf{$IV$}: The addition of new modules requires both a Morphology-mounting command and a module. If the slot-reference is empty when interpreting a module, the module is not expressed in the phenotype, except for the $C$ module, which is the very first module and needs no mounting command. When the \textit{module-reference} is a joint, an attempt to attach it to the front slot is made, regardless of the mounting command. When the \textit{module-reference} is the core-component, if its left, front, and right slots are occupied, an attempt to attach it to the back slot is made, regardless of the mounting command. If the mounting attempt is made to a slot that is occupied, the module is not expressed, while the command remains in the \textit{slot-reference}. If the newly mounted module intersects an existing one during the development, both the new module and its associated network node (if there is one) are not expressed. After mounting a new module, the \textit{module-reference} remains in the parent module, and the \textit{slot-reference} is emptied.

 \item \textbf{$V$}: The Morphology-moving commands update the \textit{module-reference} according to the slot defined by the command. If the \textit{module-reference} is a joint, any Morphology-moving command moves to the front slot.

 \item \textbf{$VI$}: The Controller-moving commands update the \textit{neuron-reference} or \textit{input-reference} according to the steps defined by the command, and is divided into two steps. The steps are illustrated by Fig~\ref{fig:fts}.

 \item \textbf{$VII$}: The Controller-changing commands apply changes to the \textit{neuron-reference} and/or \textit{input-reference}, or to the edge connecting them. Controller-changing commands act upon the input/neuron nodes at the top (latest) of the stack. If there are no input/neuron nodes yet (according to the requirements of the command), the command is not expressed.
 
 If a newly mounted module is a joint, a new neuron is created possessing a connection weight that is drawn from a random uniform distribution between $-1$ and $1$, and this neuron becomes the \textit{neuron-reference}. When a new neuron is created, this generates an edge between this neuron and the \textit{input-reference}. If there is no input yet, the neuron is stacked (oldest neuron remains as \textit{neuron-reference}). If there is a stack of inputs, the new neuron is connected to all of them; for the edges, the input on the top of the list uses the weight possessed by the neuron, while all the other inputs in the stack use their own weights; finally, the stack is partially emptied keeping only the latest neuron, which becomes the \textit{neuron-reference}.
  If a newly mounted module is a sensor, a new input is created possessing a connection weight that is drawn from a random uniform distribution between $-1$ and $1$, and this input becomes the \textit{input-reference}. When a new input is created, this generates an edge between this input and the \textit{neuron-reference}. If there is no neuron yet, the input is stacked (the oldest input remains as \textit{input-reference}). If there is a stack of neurons, the new input is connected to all of them; for the edges, the neuron on the top of the list uses the weight possessed by the input, while all the other neurons in the stack use their own weights; finally, the stack is partially emptied keeping only the latest input, which becomes the \textit{input-reference}.

For every new edge created from an input to a neuron, the edge is attributed a serial ID within the neuron. Analogously, for every new edge created from a neuron to an input, the edge is attributed a serial ID within the input. 
 
\end{itemize}
An example of late-development is illustrated in Fig~\ref{fig:decoding}.
 
\subsection{Initialization}
\label{sec:genoinit}

\begin{figure}[t]
    \begin{center}
      \includegraphics[width=5.2in]{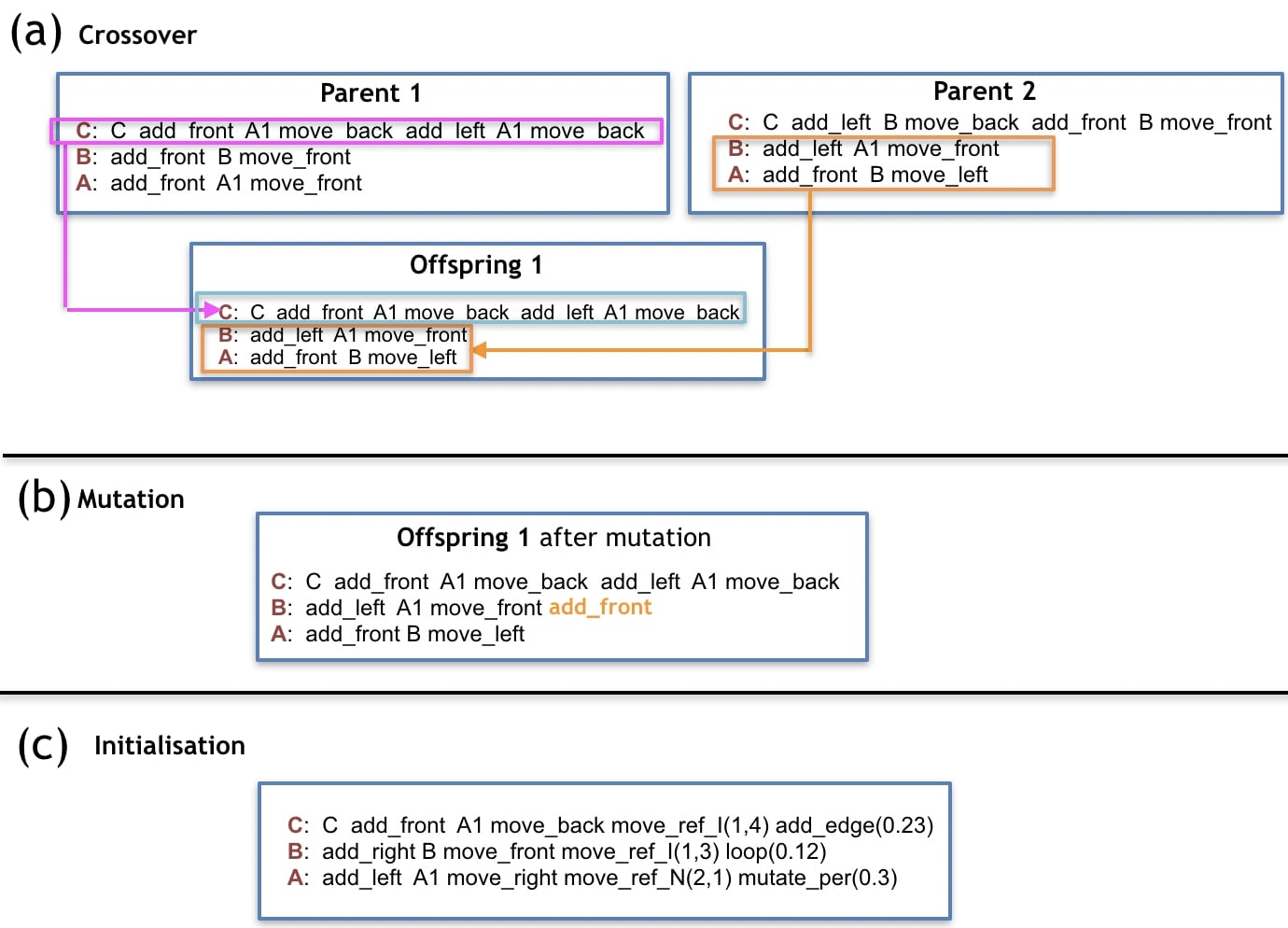}  
       \end{center}
           \caption{a) and b) are examples of reproduction operators, and c) is an example of initialization using only $1$ group of symbols for all cases of rules.}
        \label{fig:operators}
\end{figure}

To initialize a genotype in the \textit{Baseline}, for each production-rule, exactly one symbol is drawn uniformly random from each of the following categories in this order: Controller-moving commands, Controller-Changing commands, Morphology-mounting commands, Modules, Morphology-moving commands. This process is repeated $s$ times, being $s$ sampled from a uniform random distribution ranging from $1$ to $e$. This means that each rule can end up with $1$ or maximally $e$ sequential groups of five symbols. The symbol C is reserved to be added exclusively (and surely) at the beginning of the production rule C. (Fig~\ref{fig:operators}.c) 

In the case of the \textit{Plasticoding}, the initialization of the production-rules is exactly the same as in \textit{Baseline}, with the additional initialization of the regulation clauses. Each regulation clause is initialized by selecting $z$ random terms, each term selected from all the available environmental states. $z$ can assume discrete values from $1$ to $u$, each with equal probability (the same term can be sampled multiple times). Each term is compared to a value chosen randomly between \textit{True} or \textit{False} with equal probability, and if $z$ is above $1$, the terms are connected by operator(s) chosen randomly between \textit{and} or \textit{or} with equal probability.

\subsection{Crossover}
\label{sec:crossover}

For the \textit{Baseline}, the crossovers are performed by selecting individual production-rules, each represented by one replaceable  symbols. The selection is performed uniformly at random from the parents (Fig~\ref{fig:operators}.a). 

In \textit{Plasticoding}, the process is similar to \textit{Baseline}, in the sense that groups of production-rules are selected individually, together with their regulation clauses, where the grouping of production-rules is defined so that each group must be associated to the same replaceable symbol.

\subsection{Mutation}
\label{sec:mutation}

  There is an equal chance of a mutation happening to any production-rule of a grammar. For the production-rule chosen to be mutated, there is an equal chance of adding/deleting/swapping one random symbol from a random production-rule/position, or apply a change to its regulation clause. (Fig~\ref{fig:operators}.b). 
  
 All symbols have the same chance of being removed or swapped. As for the addition of symbols, all categories have an equal chance of being chosen to provide a symbol, and every symbol of the category also has an equal chance of being chosen. An exception is made to $C$ to ensure that a robot has one and only one core-component. This way, the symbol $C$ can not be added to any other production rules, neither removed or moved from the production rule of $C$. The operations adding/deleting/swapping have an equal chance to happen.
 
 In the case of changing a regulation clause (for \textit{Plasticoding}), there is an equal chance of adding/removing a term from the clause (by still ensuring that the number of terms is between $[1,u]$, or flipping a state of a term from \textit{True} to \textit{False} (or the inverse), or to flip an operator from \textit{and} to \textit{or} (or the inverse).

\subsection{Evolution}
\label{sec:evolution}
   We are using overlapping generations with a population size $\mu=100$. In each generation, $\lambda=100$ offspring are produced by selecting $100$ pairs of parents through binary tournaments (with replacement) and creating one child per pair by crossover and mutation. From the resulting set of $\mu$ parents plus $\lambda$ offspring, $100$ individuals are selected for the next generation, also using binary tournaments. The evolutionary process is stopped after $200$ generations, thus all together we perform $20000$ fitness evaluations per run. For each environmental scenario, the experiment was repeated $20$ times independently. A summary of the parameters for the evolutionary algorithm is provided in Table~\ref{tab:params}.

 \begin{table}[!ht]
 
\centering
\caption{  
{\bf Parameters for the evolutionary algorithm.}}
\begin{tabular}{|l|l|}
  \hline

  Population size & $100$  \\ \hline
   Offspring size & $100$  \\ \hline
  Number of generations & $200$  \\ \hline
    Mutation probability & $80$\%  \\ \hline
  Crossover probability & $80$\%  \\ \hline
   Rewriting iterations $k$  & $3$  \\ \hline
  Maximum number of groups of symbols $e$ & $4$  \\ \hline
   Connections of the network range from & $-1$ to $1$  \\ \hline
   Oscillator parameters range from & $1$ to $10$  \\ \hline
  Maximum amount of modules $m$ & $15$  \\ \hline
  Experiment repetitions & $20$  \\ \hline

\end{tabular}

\label{tab:params}
\label{table2}
 
\end{table}

 \section{Experimental setup}

 We carried out two sets of experiments using the same experimental setup, except for the encoding method. The first set of experiments used the \textit{Baseline} encoding, while the second set used the \textit{Plasticoding} encoding.
 
Our experiments were realized using a simulator called Gazebo, interfaced through a robot framework called Revolve~\citep{hupkes2018revolve}.
The code needed to reproduce our experiments and analysis in available on GitHub\footnote{\url{https://github.com/ci-group/revolve/tree/ae99a0985765997e5e5b557bc677f4cc1bc84c99/experiments/plasticoding_frontiers2020}}.

\subsection{Environmental conditions}

\begin{figure}[t]
    \begin{center}
         \includegraphics[width=4in]{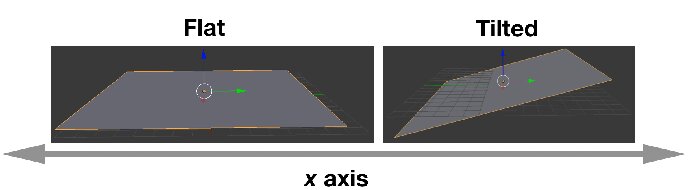}
    \caption{Environmental conditions: Flat and Tilted.}
        \label{fig:environments}
    \end{center}
    \end{figure}

\begin{figure}[t]
    \begin{center}
         \includegraphics[width=3.5in]{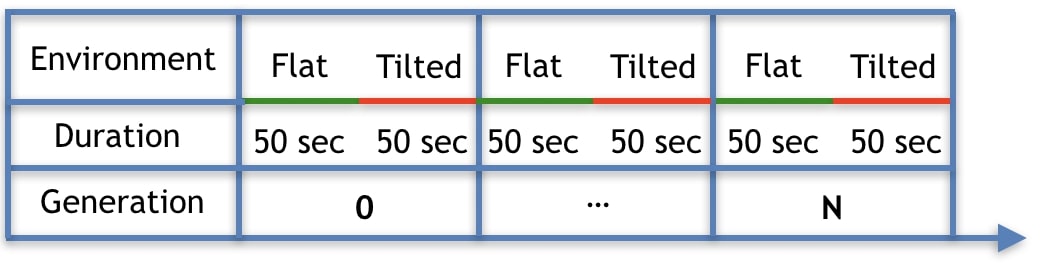}
    \caption{Seasonal environmental condition.}
        \label{fig:seasons_setup}
    \end{center}
    \end{figure}

We experimented with two different environmental conditions, which are a) Flat environmental condition: it is a plane flat floor; b) Tilted environmental condition: it is a plane floor tilted in $5$ degrees. These conditions are depicted in Fig~\ref{fig:environments}. In the experiments, these conditions were combined to create a seasonal environmental condition. This way, robots live each part of their lifetime in one different environmental condition. They spend their first $50$ seconds of lifetime in the Flat environmental condition, and after that they spend $50$ more seconds in the Tilted environmental condition (Fig~\ref{fig:seasons_setup}). Note that during morphogenesis robots are in the Flat environmental condition, and later on during their life, have the chance to experience the Tilted environmental condition regardless their performance in the previous condition. 
    
 \subsection{Fitness function }
 \label{sec:fitness}

For each environmental condition independently, Flat and Tilted, the fitness function is defined by Eq.~\eqref{eq:hill}.

\begin{equation}
\begin{gathered}
        f_1 = \begin{cases}
       s_x &\text{if } s_x > 0  \\
        \frac{s_x}{10} & \text{if } s_x < 0\\
         -0.1  & \text{if } s_x = 0 
      \end{cases}                              
  \end{gathered}
\label{eq:hill}
\end{equation}

\noindent where $s_x$ is the speed of the robot as defined by Eq.~\eqref{eq:speed}. This function measures the speed of the robots only in the $x$ axis, so to discourage robots to exploit locomotion in the $y$ axis, avoiding the proposed challenge of climbing the Tilted environmental condition. Additionally, there are two penalties to try to escape local optima observed in preliminary tests. The first penalty is the division by $10$ used when the speed is negative, which aims at preventing that a ``safe strategy'' be much more beneficial than falling completely down the hill. This ``safe strategy'' is characterized by trying to avoid to fall too far from the starting point (due the effect of gravity), but without really climbing. The second penalty is the constant $-0.1$ used when speed is zero, which aims at disincentivizing robots that do not develop joints (and thus can not move) so to  avoid the risk of falling. 

Because robots are evaluated in multiple environmental conditions, we treat this problem as multi-objective, where the fitness of each environmental condition represents one of the objectives. Notably, this setup in which robots need to perform well in different environmental conditions can be seen as robots having to perform well different tasks. To obtain the final fitness, which represents the fitness in the seasonal environmental condition, we consolidate the two fitness values into a single measure. The consolidation of these objectives into the final fitness is defined by Eq.~\eqref{eq:cons}.
     
       \begin{eqnarray}
    \label{eq:cons}
    \begin{aligned}
      f_c = \sum_{i=1}^{n}d_i
    \end{aligned} 
    \end{eqnarray}

\noindent where $d_i$ is the number of solutions in the population that solution $i$ dominates, given that a solution only dominates another solution if it is better than this in at least one objective and not worse in any objective.

 \subsection{Robot Descriptors}
 For quantitatively assessing morphological, control, and behavioral properties of the robots, we utilized a set of descriptors.\\
  
 \subsubsection{Behavioral descriptors}

 
\textbf{1. Speed}: Describes the speed (cm/s) of the robot along the $x$ axis as defined by Eq.~\eqref{eq:speed}.

    \begin{eqnarray}
    \label{eq:speed}
    \begin{aligned}
      s_x = \frac{e_x - b_x}{t}
    \end{aligned} 
    \end{eqnarray}

\noindent where $b_x$ is $x$ coordinate of the robot's center of mass at the beginning of the simulation, $e_x$ is $x$ coordinate of the robot's center of mass at the end of the simulation, and $t$ is the duration of the simulation.

\textbf{2. Balance}:  We use the rotation of the head in the $x$--$y$ plane to define the balance of the robot. In general, the rotation of an object can be described in the dimensions roll, pitch, and yaw. We consider the pitch and roll of the robot head, expressed in degrees between 0 and 180 (because we do not care if the rotation is clockwise or anti-clockwise). Perfect Balance belongs to both pitch and roll being equal zero, so that the higher the Balance, the less rotated the head is. Formally, Balance is defined by Eq. (\ref{eq:bal}).
    
    \begin{eqnarray}
    \label{eq:bal}
    \begin{aligned}
      B = 1 - \frac{r + p}{t * 180 * 2}
    \end{aligned} 
    \end{eqnarray}
    \noindent where $r = \sum_{i=1}^{t}\mid r_i\mid$, representing the roll rotation accumulated over time, $p = \sum_{i=1}^{t}\mid p_i\mid$, representing the pitch rotation accumulated over time,  and $t$ is the duration of the simulation.\\
    

\subsubsection{Morphological Descriptors}

\begin{enumerate}
 
   \item \textbf{Size}: Total number $S$ of modules in the morphology.

   \item \textbf{Sensors} Accounts for touch sensors in the morphology (Fig.~\ref{fig:sensors}). It is defined with Eq.~\eqref{eq:sensors}:

\begin{equation}
\begin{gathered}
        C = \begin{cases}
      \frac{c}{c_{max}}, &\text{if } c_{max}>0  \\
                        0 &\text{otherwise}
                    \end{cases}                              \end{gathered}
\label{eq:sensors}
\end{equation}
where $c$ is the number of sensors and $c_{max}$ is the number of slots in the morphology that are not connected to other types of module.

\begin{figure}
\begin{center}
\includegraphics[width=1.5in]{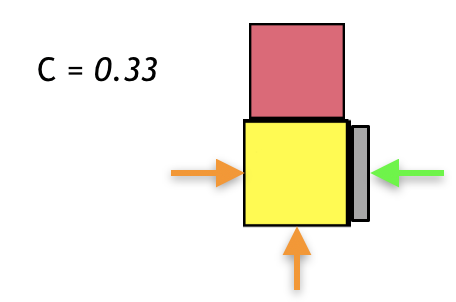}
\vspace{-0.5cm}
\caption{Example morphology containing a single sensor (green arrow), and two more free slots (orange arrows).}
\label{fig:sensors}
\end{center}
\vspace{-0.3cm}
\end{figure}
 
\end{enumerate}

\subsubsection{Controller Descriptors}

\begin{enumerate}
 
   \item \textbf{Sensors Reach}: Describes how the inputs from the sensors are connected to the oscillators of the controller. The higher this number, the more motors each sensor is sending data to in average (Fig.~\ref{inputs_reach}). It is defined with Eq. (\ref{eq:inputs_reach}):

\begin{equation}
\label{eq:inputs_reach}
\begin{aligned}
D_{s} = Md(R_s) \ \ \ \ \ \ \ \ \ \ \ \ \ \ \ \ \ \ \ \ \ \ 
           \\
R_s = \Bigg\{ r_s \ | \ r_s = \frac{c_s}{n(L)} \ \forall \ s \in S \Bigg\}
\end{aligned} 
\end{equation}
\noindent where $R_s$ is a set of ratios, while $c_s$ is the number of connections of the input $s$, $S$ is the set of all inputs in the controller, and $n(L)$ is the number of oscillators in the controller.
 
 \begin{figure}[!htb]
\centerline{\includegraphics[scale=0.2]{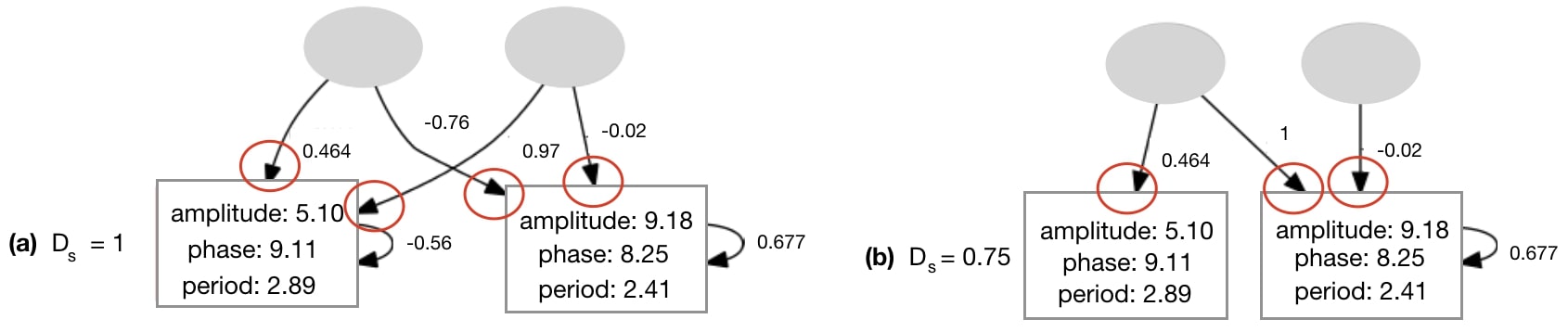}}
\caption{Controller (a) has inputs connected to more oscillators than controller (b) in average.}
\label{inputs_reach}
\end{figure}

   \item \textbf{Recurrence}: Describes the proportion of oscillators in the controller that have a recurrent connection, i.e., memory (Fig.~\ref{recurrence}). It is defined with Eq. (\ref{eq:recurrence}): 

\begin{equation}
\label{eq:recurrence}
\begin{aligned}
           D_r = \frac{r(L)}{n(L)}
\end{aligned} 
\end{equation}
\noindent where $n(L)$ is the number of oscillators of the controller, and $r(L)$ is the number of oscillators that have a recurrent connection.

 \begin{figure}[!htb]
\centerline{\includegraphics[scale=0.2]{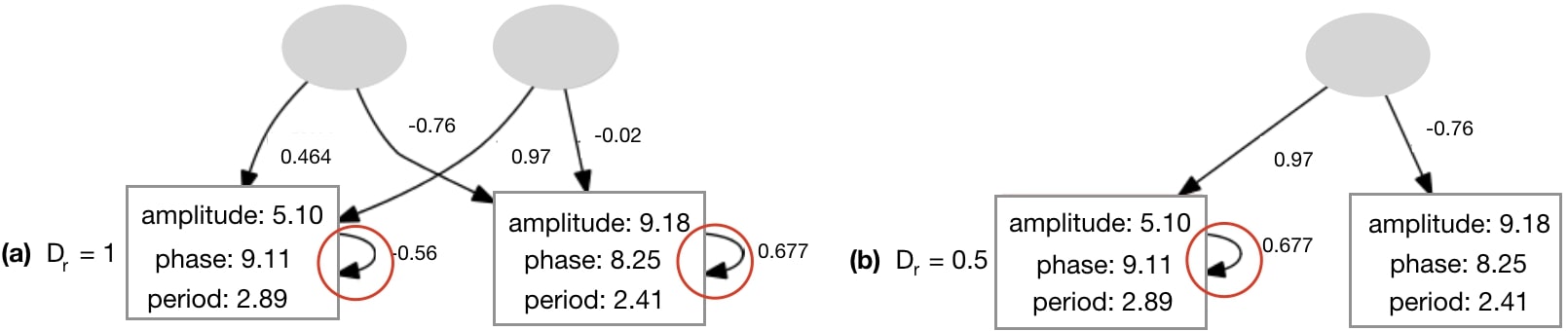}}
\caption{Controller (a) has more recurrence than controller (b).}
\label{recurrence}
\end{figure}

\end{enumerate}

A complete search space analysis of the utilized robot framework and its descriptors is available in~\citep{mirassearch2017, 2miras2018}, demonstrating the capacity of these descriptors to capture relevant robot properties, and proving that this search space allows high levels of diversity.

\section{Results and Discussion}

When robots have to cope with multiple environmental conditions while disposing of one same morphology and controller, and thus behavior, naturally a trade-off occurs. Because of the need to adapt to different environmental conditions, in at least one of the environmental conditions they might adapt worse than if they had evolved in that same static environmental condition. In this case, for the seasonal environmental condition, where both Flat and Tilted had to be faced by the population, the Tilted season exerted a higher selection pressure. This way, robots acquired the same traits as robots that had evolved in a static Tilted environmental condition. One probable reason for this is that, as demonstrated in another study~\citep{miras2019impact}, robots evolved in an inclined environment can still perform the task in the a flat environment, but fail badly when it is the other way around, showing that the pressure of an inclined environment leads to more generalist strategies for locomotion.
 
\begin{figure}[t]
\centering
    \includegraphics[width=2in]{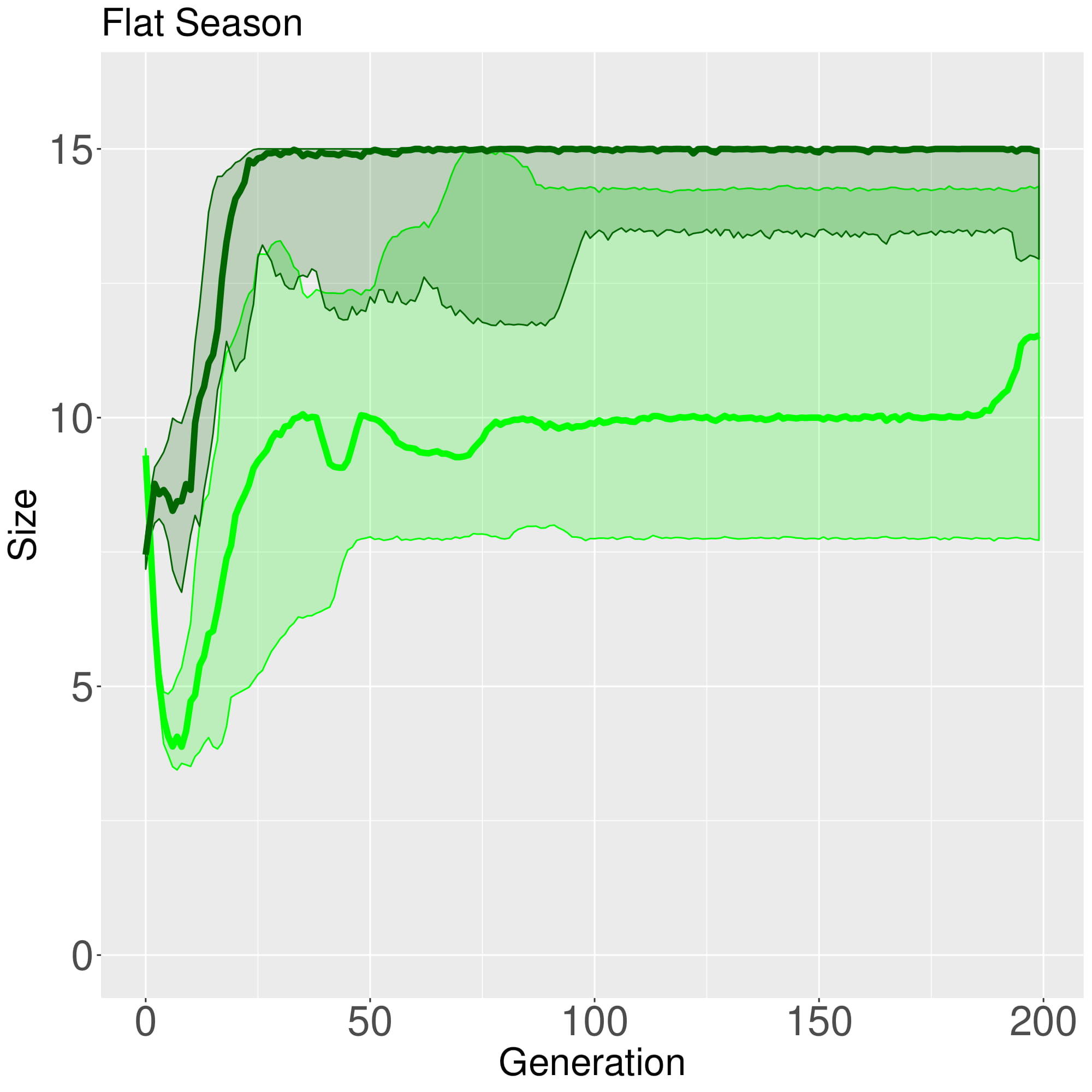}
      \includegraphics[width=2in]{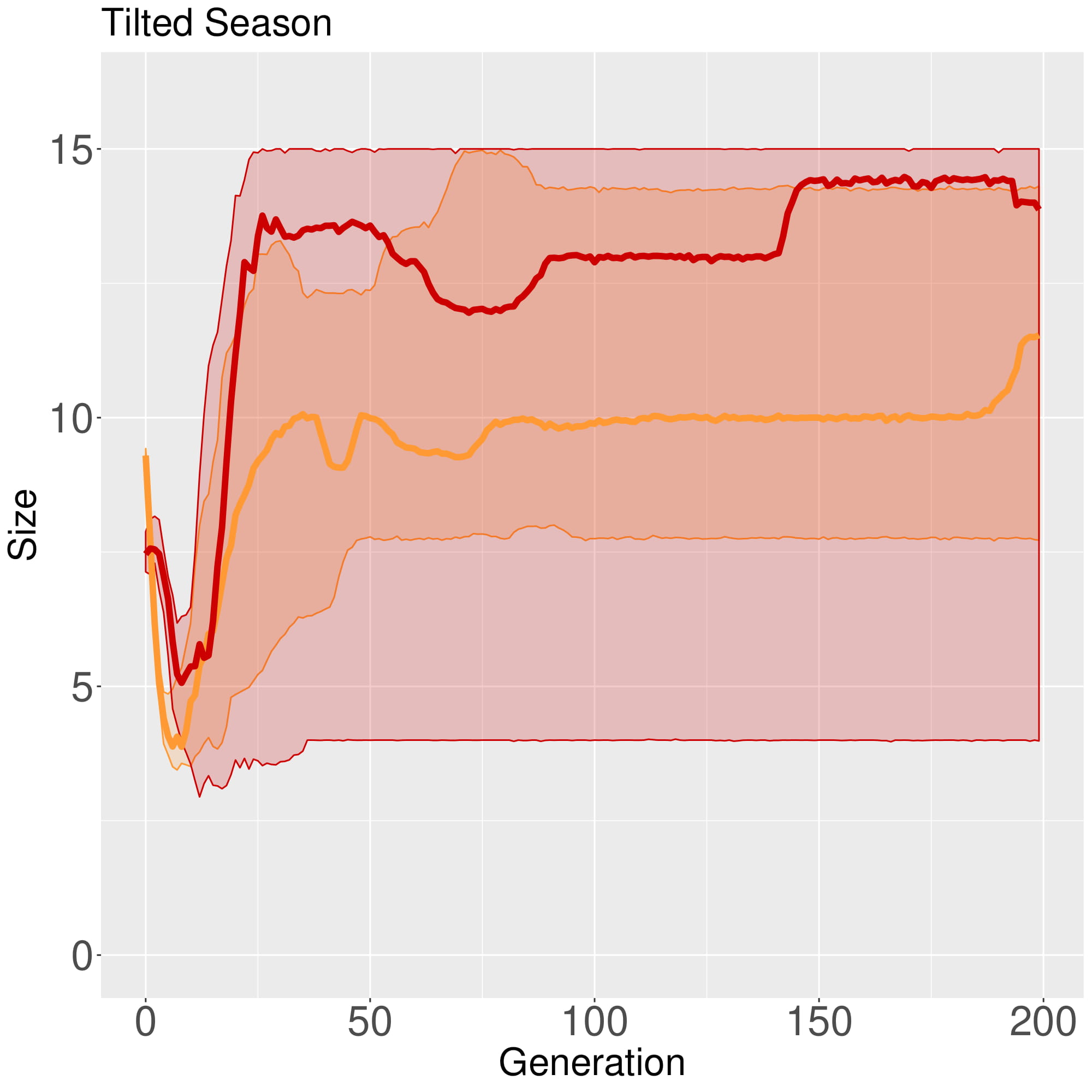}
    \includegraphics[width=1.1in]{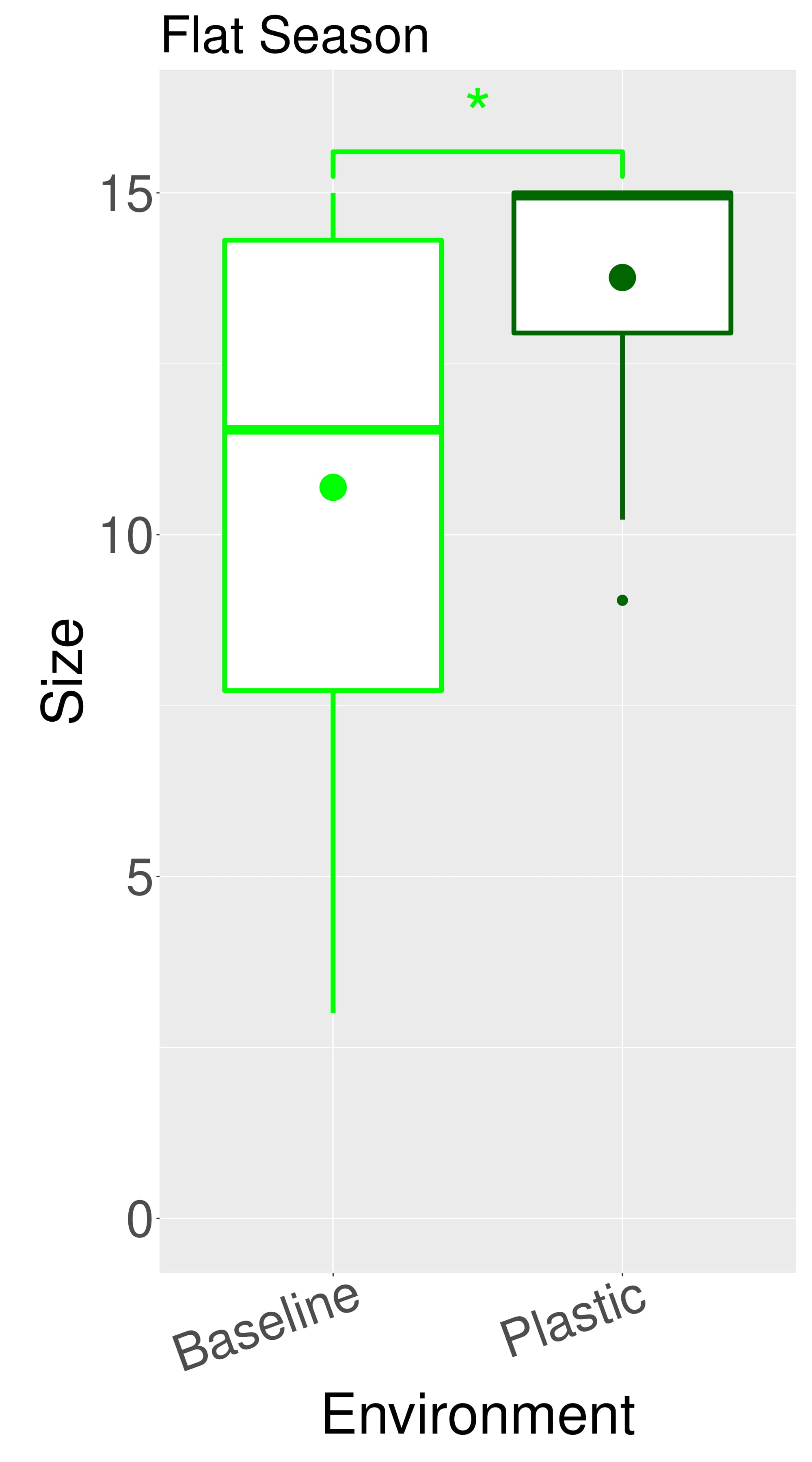}
    \includegraphics[width=1.1in]{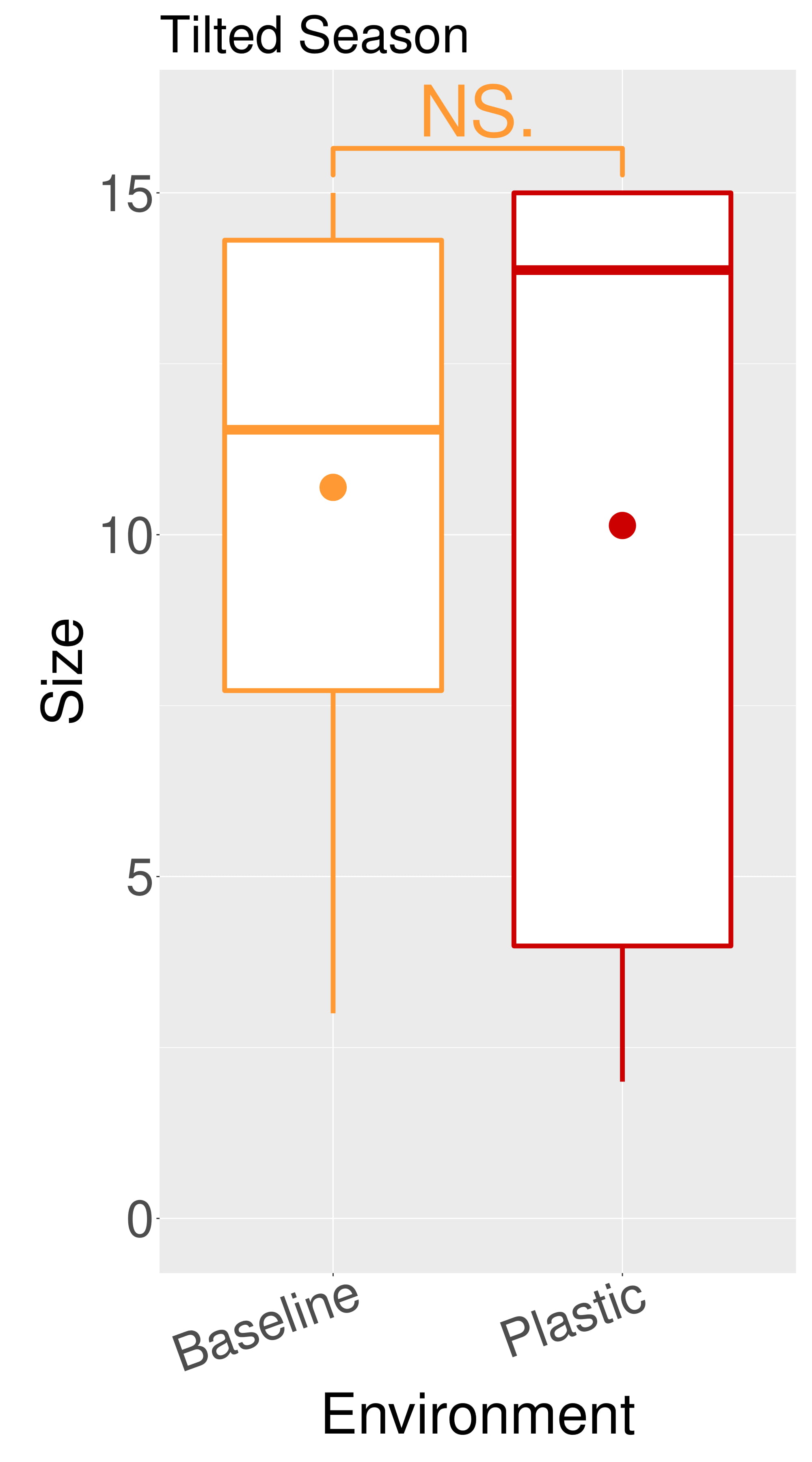}
        \\
    \includegraphics[width=2in]{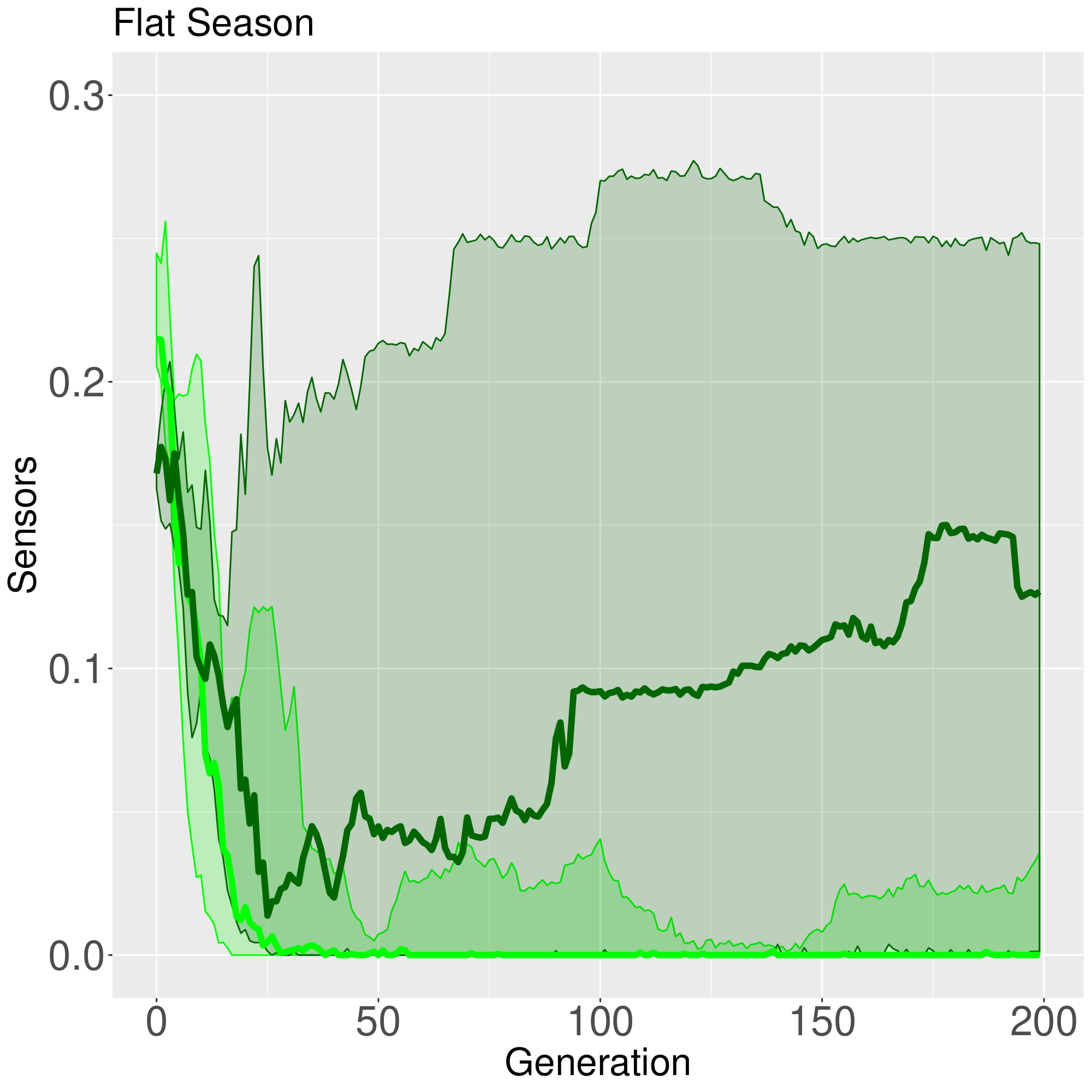}
      \includegraphics[width=2in]{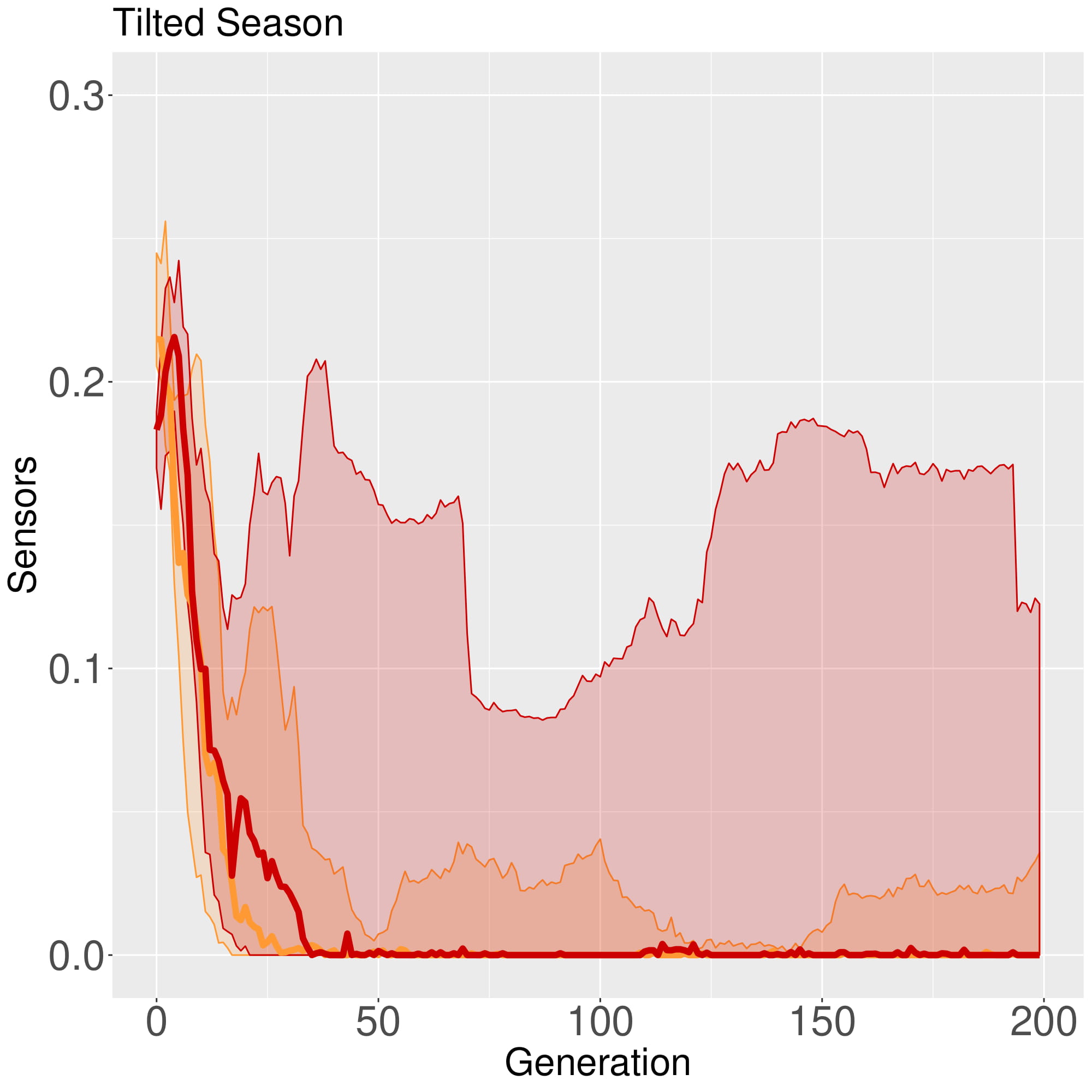}
    \includegraphics[width=1.1in]{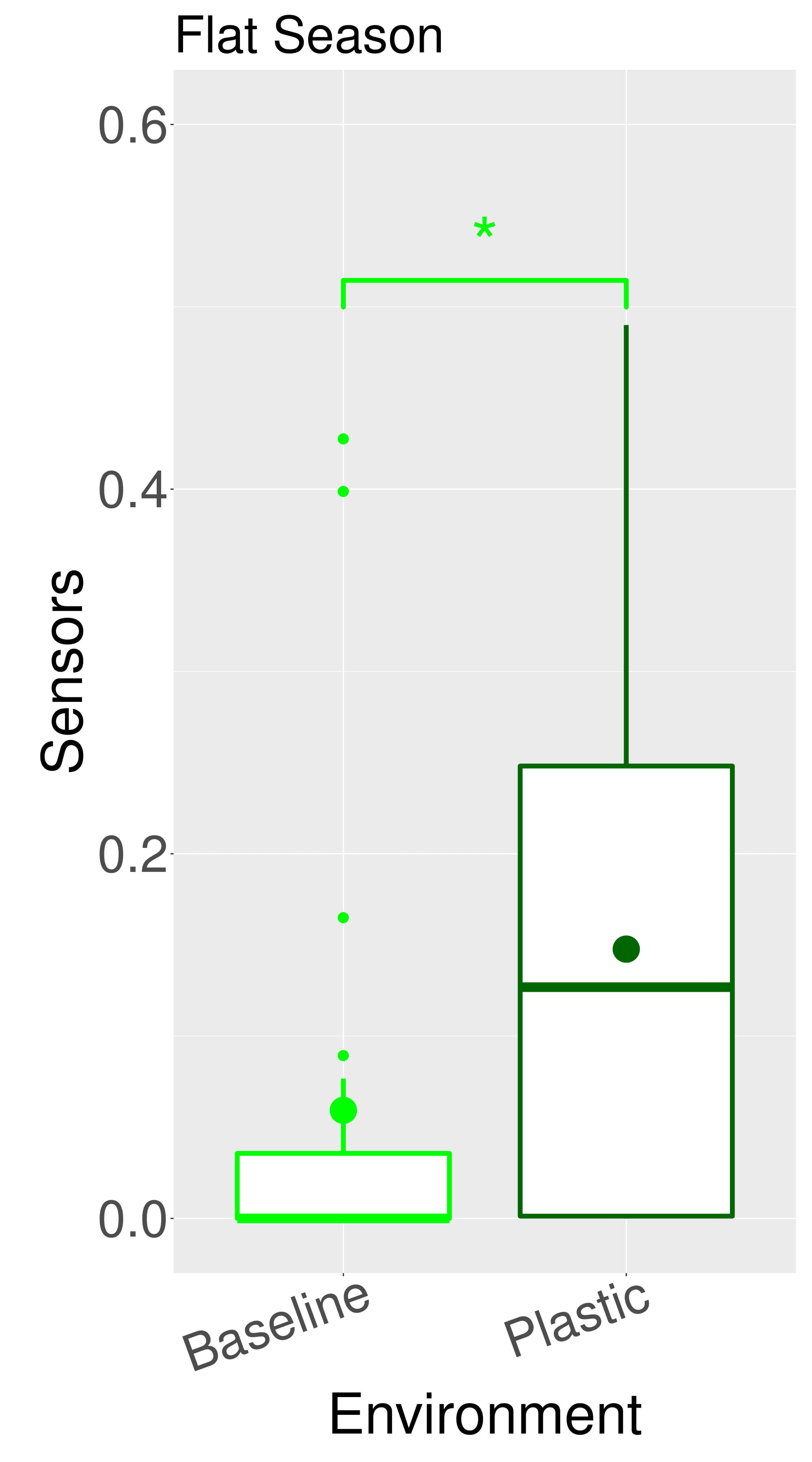}
    \includegraphics[width=1.1in]{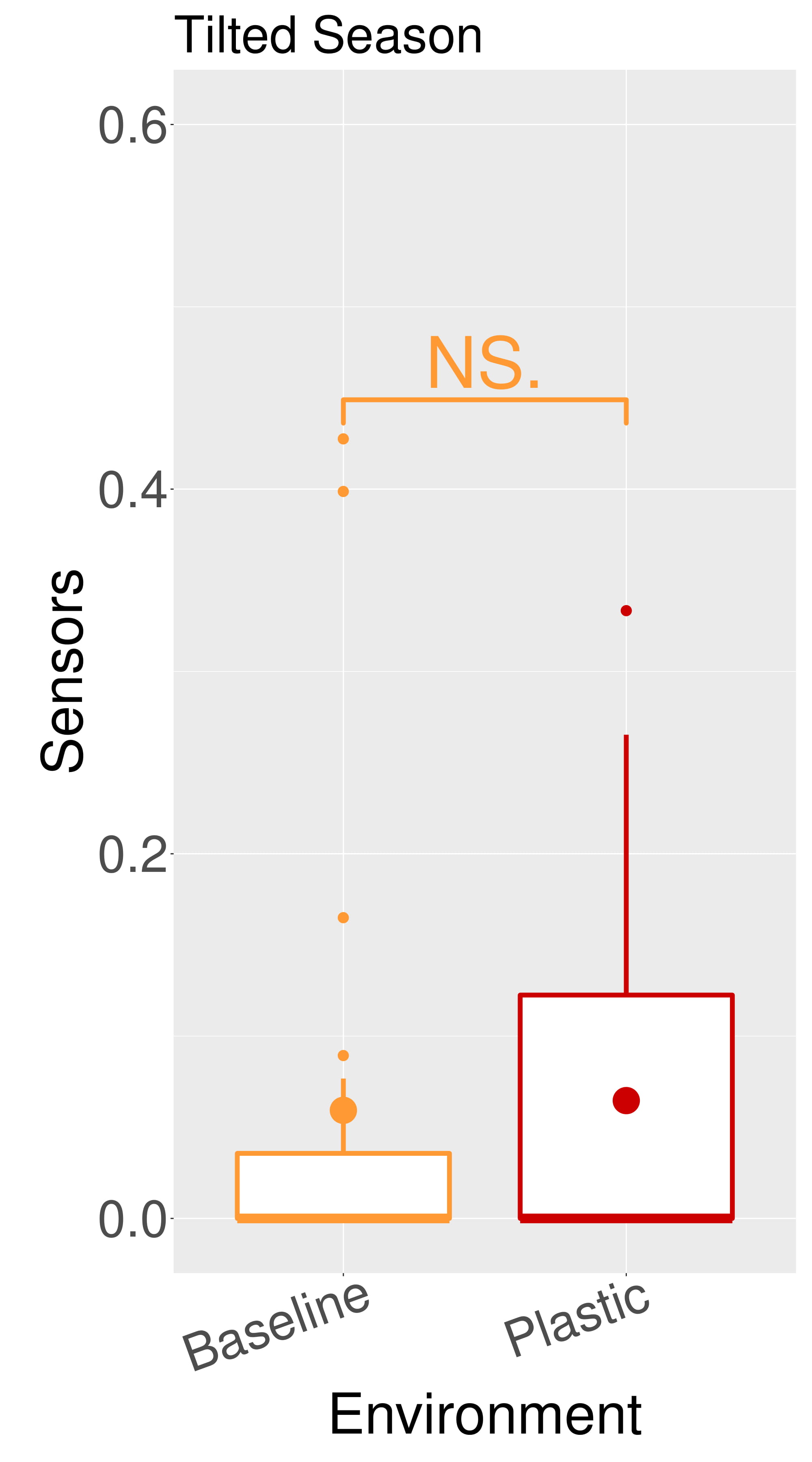}
    \\
 
    \begin{center}
     \includegraphics[width=4.5in]{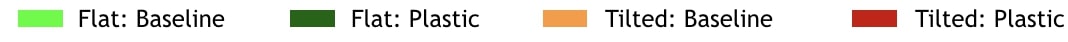}
        \end{center}
    \caption{Comparison of morphological properties in different environmental conditions. \textbf{Line plots} show the progression of the means of the population (quartiles over all runs), while \textbf{boxplots} show the distribution of the means in the final generation. Significance levels for the Wilcoxon tests in the boxplots are $* < 0.05$, $** < 0.01$, $*** < 0.001$. Note that, naturally, the values of Baseline are the same in Flat or Tilted.}
    \label{fig:morphologies_gens_seasonal}
 
\end{figure}

\begin{figure}[t]
\centering
    \includegraphics[width=2in]{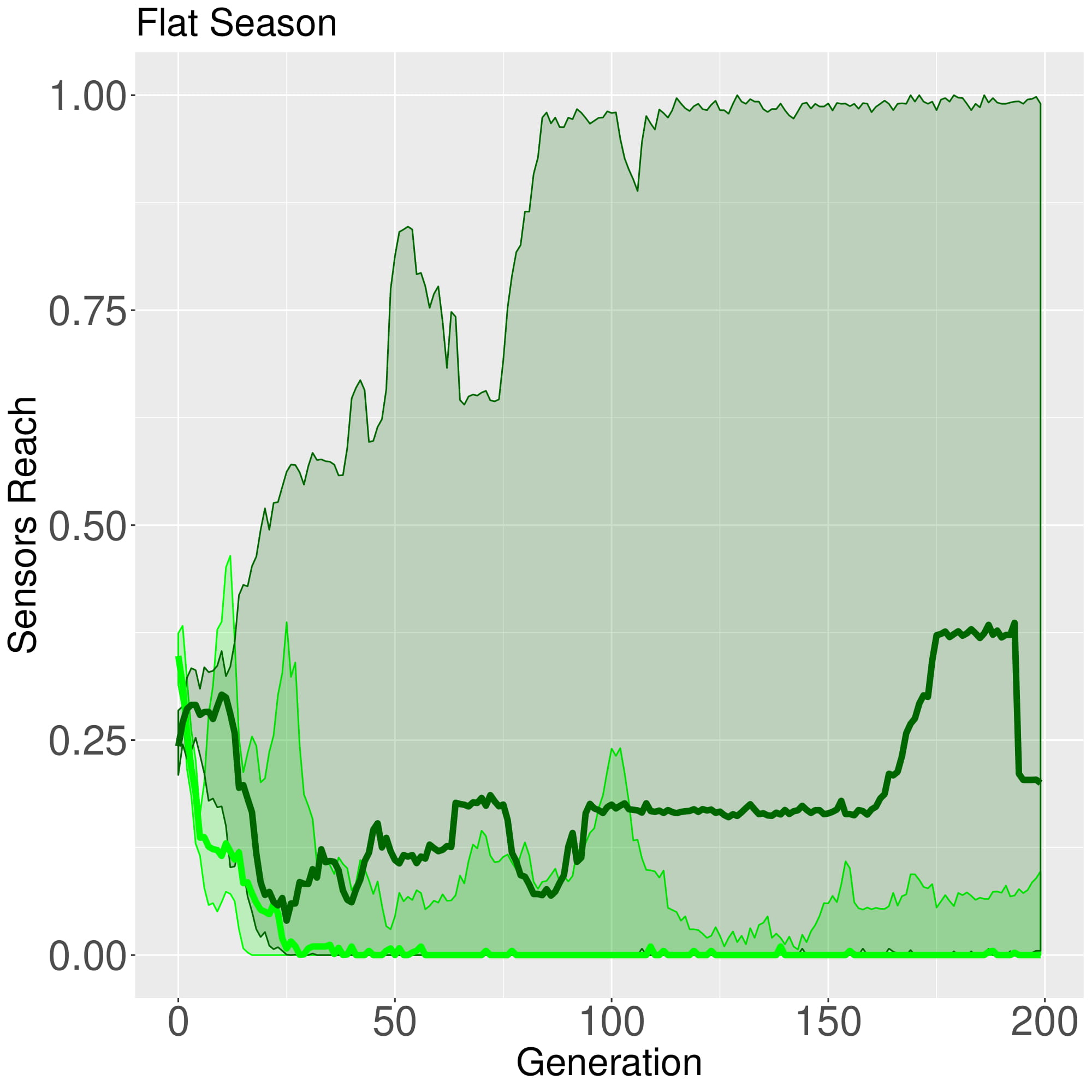}
      \includegraphics[width=2in]{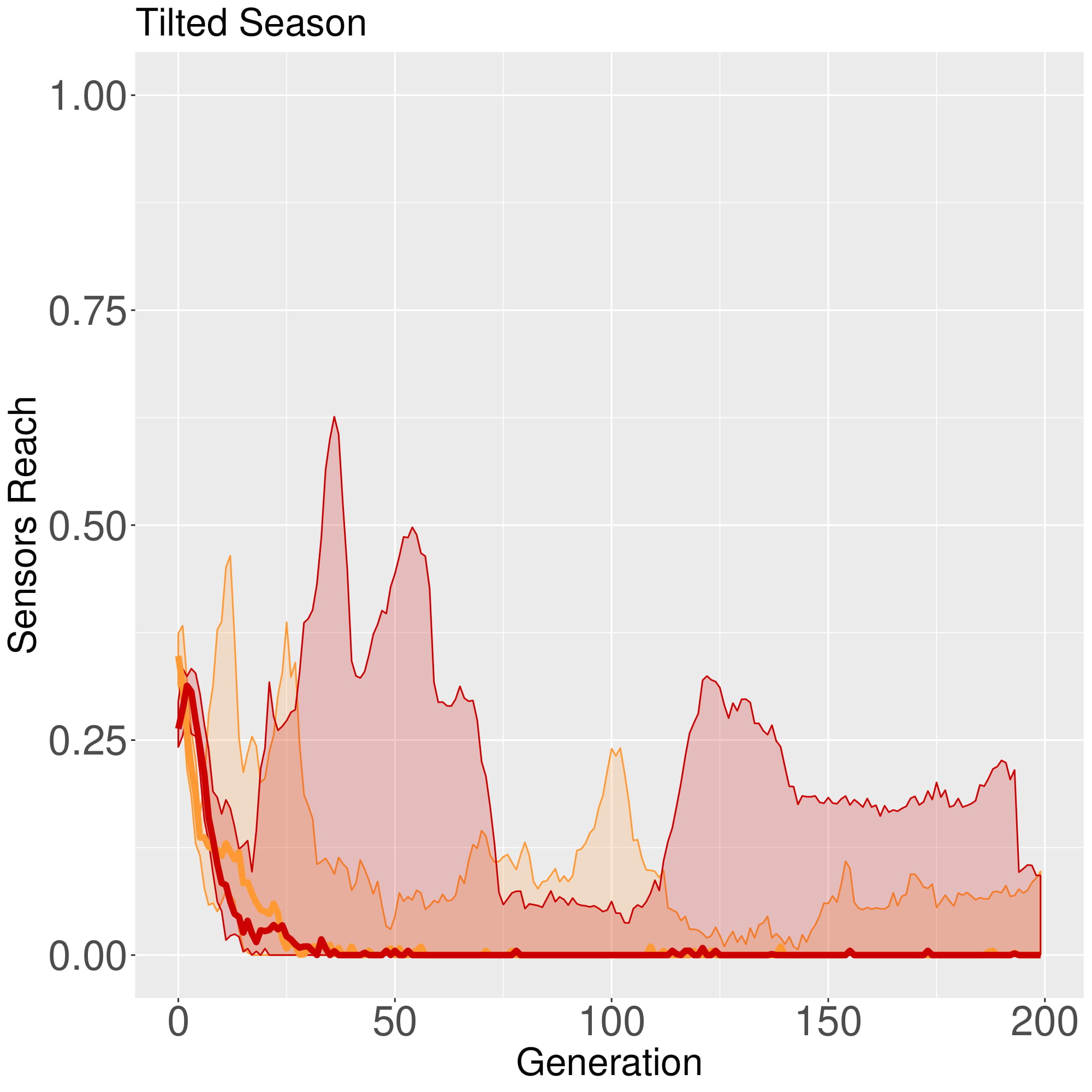}
    \includegraphics[width=1.1in]{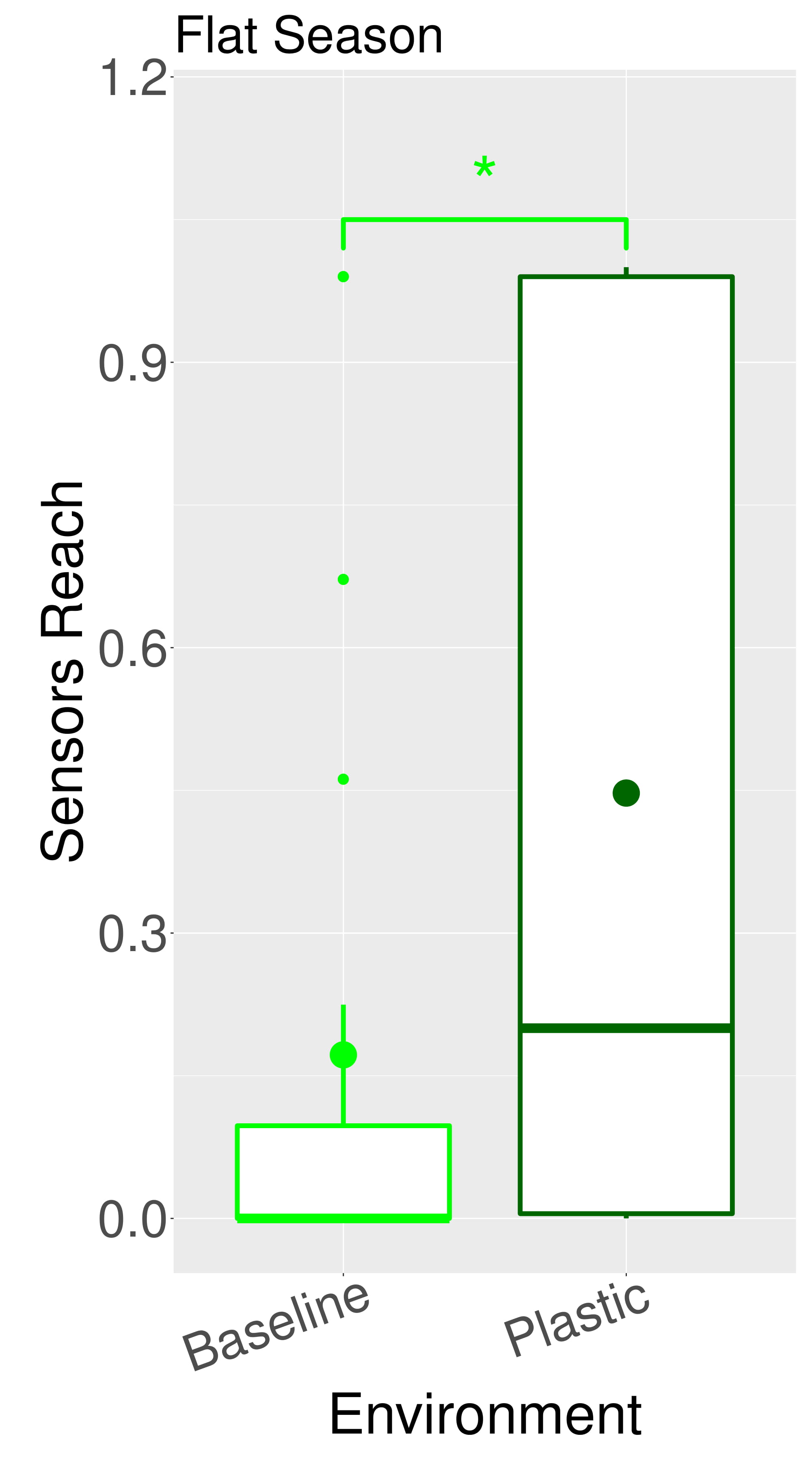}
    \includegraphics[width=1.1in]{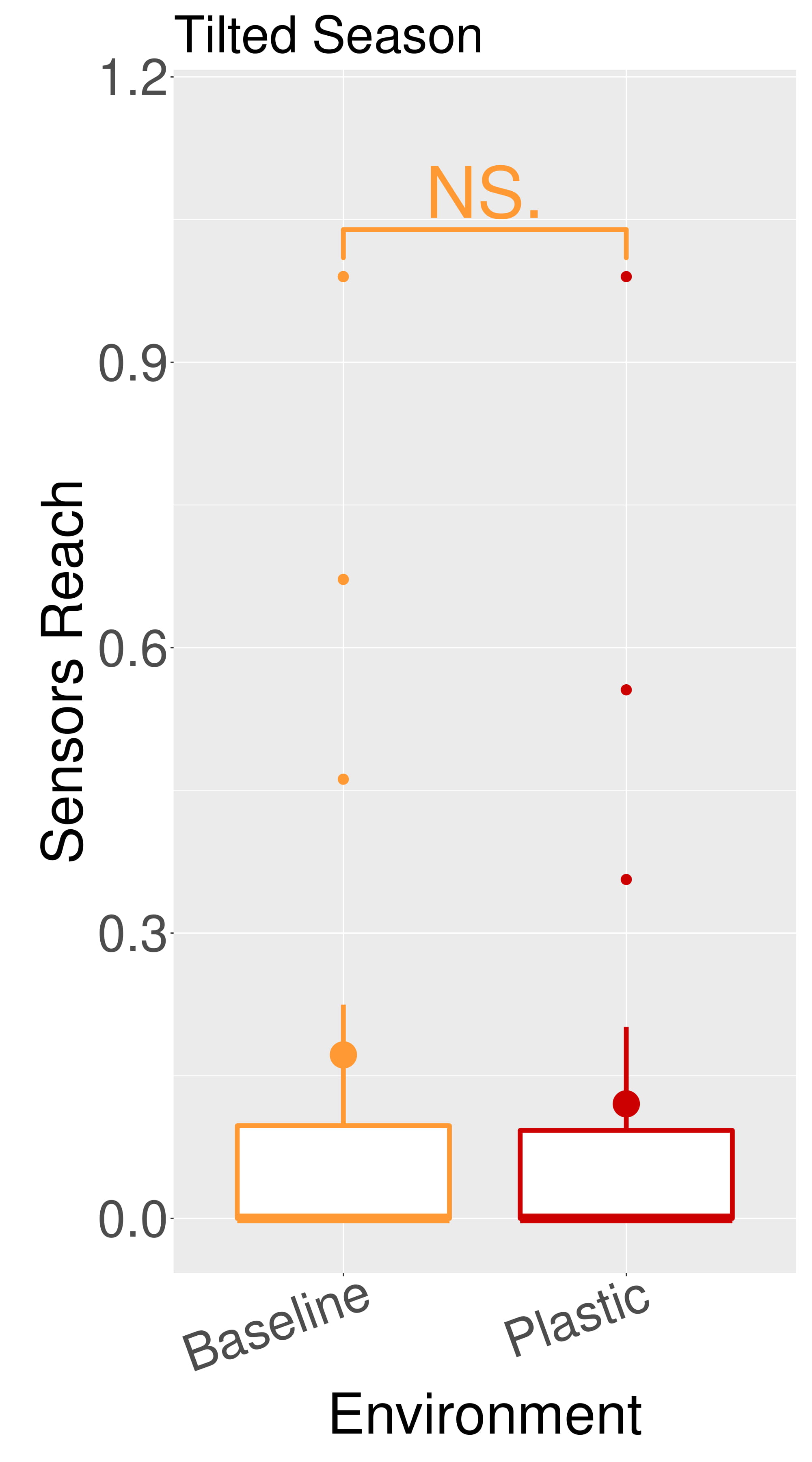}
        \\
    \includegraphics[width=2in]{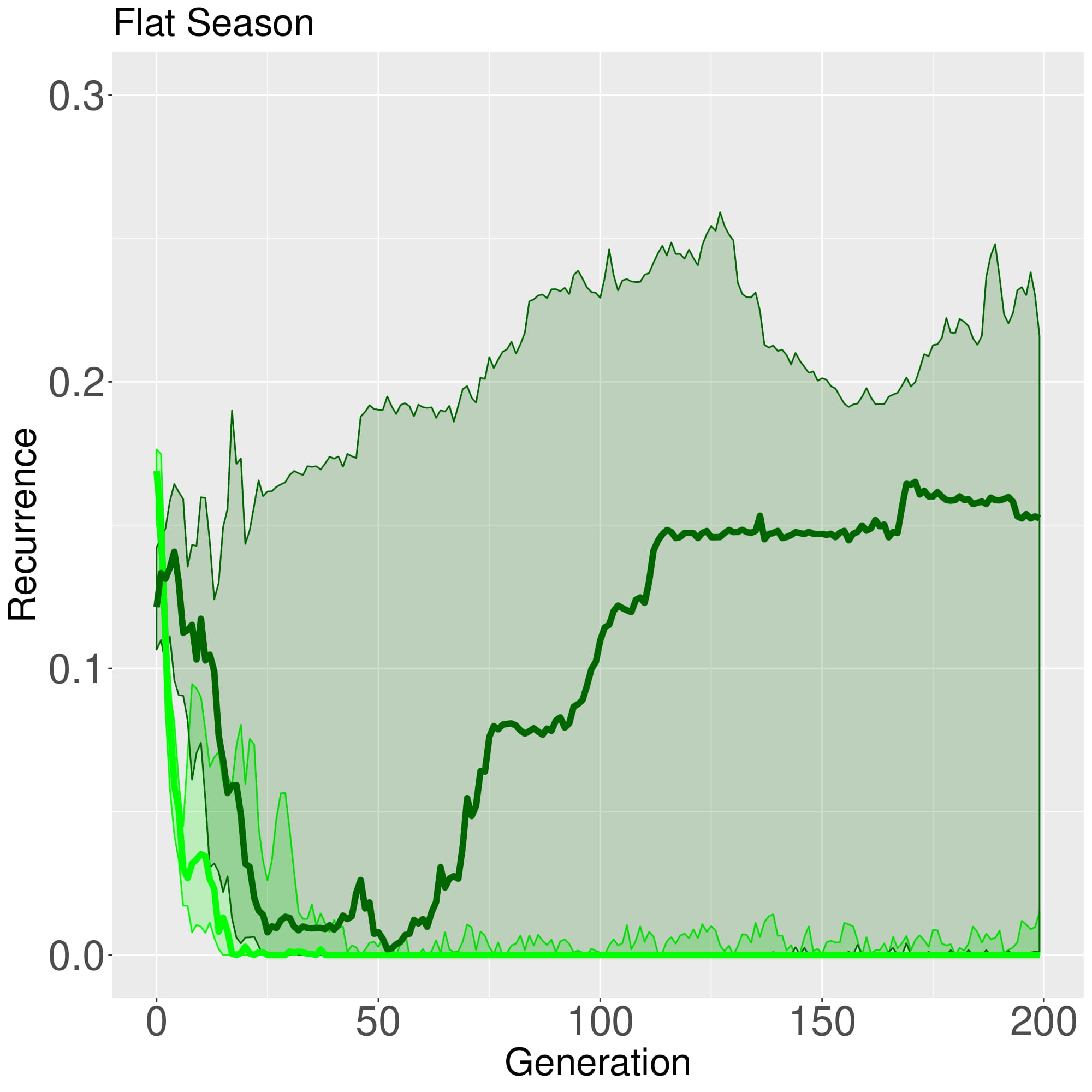}
      \includegraphics[width=2in]{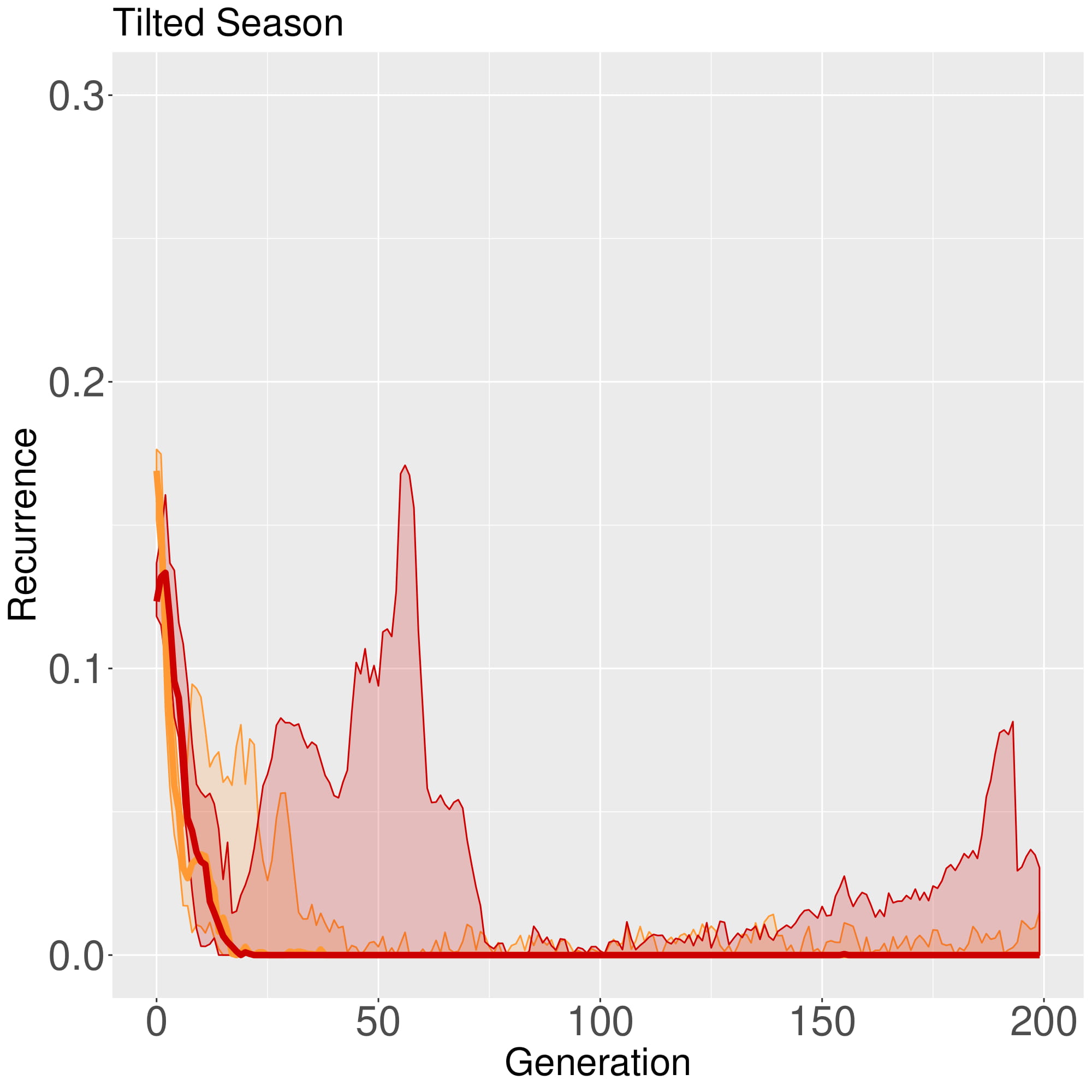}
    \includegraphics[width=1.1in]{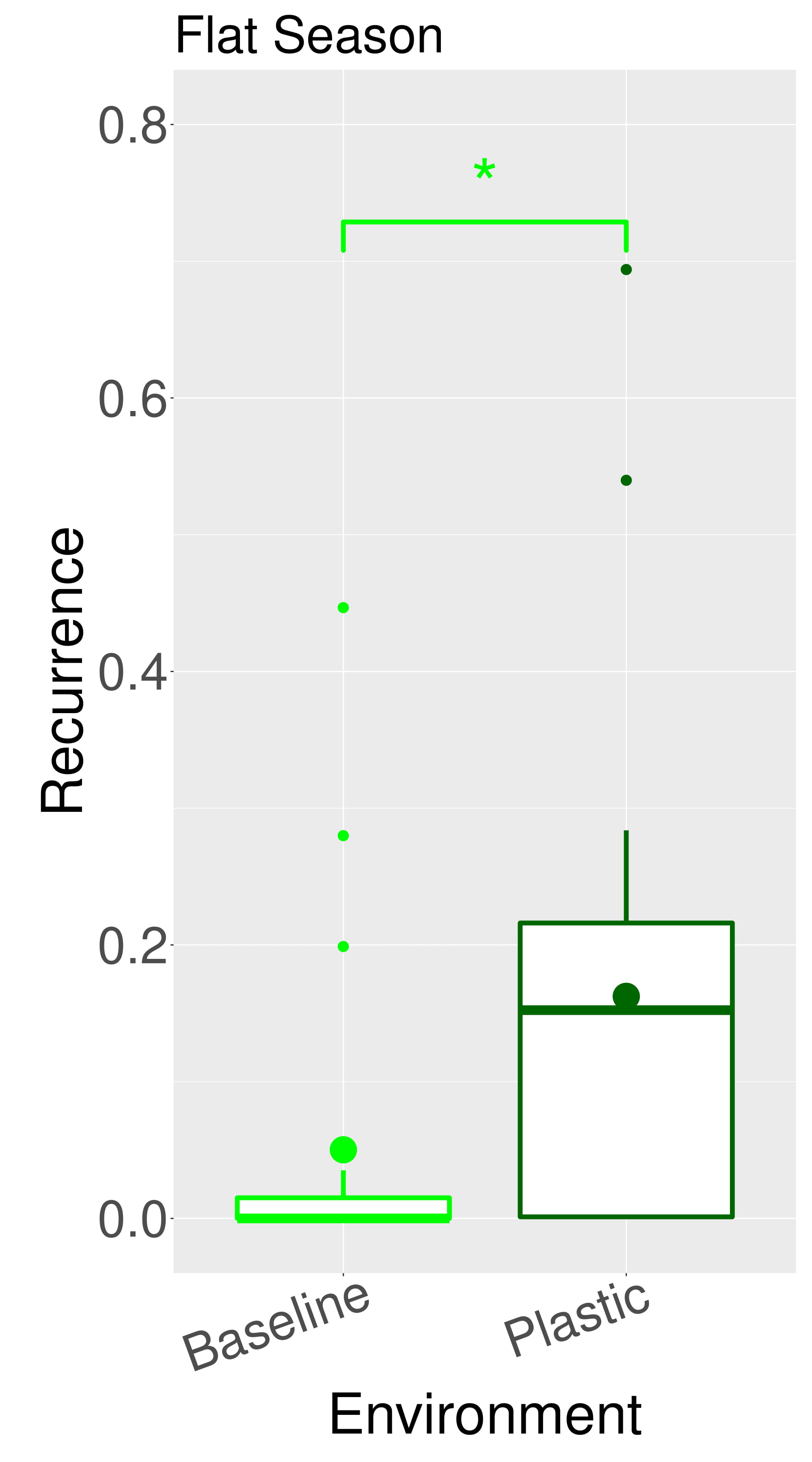}
    \includegraphics[width=1.1in]{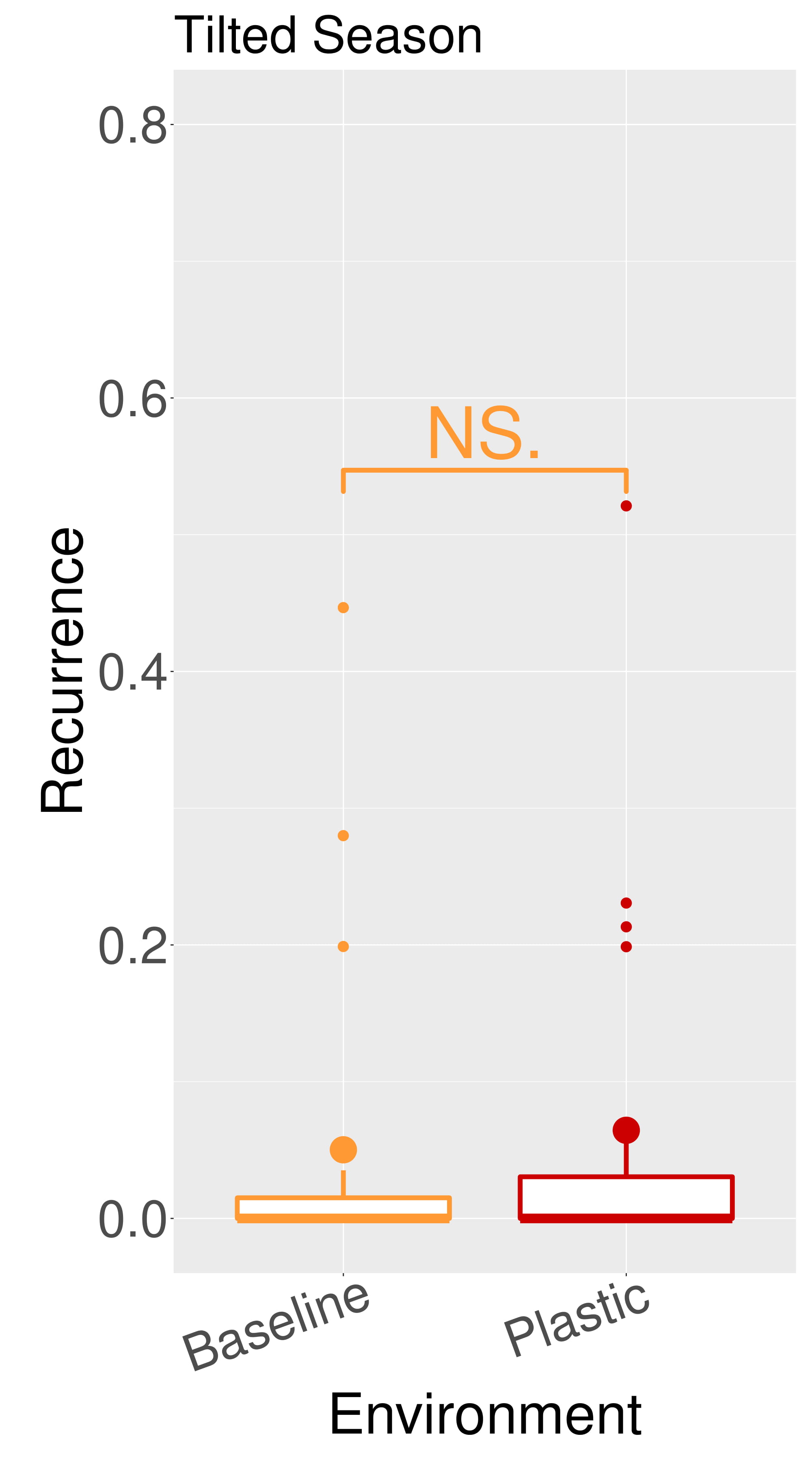}
    \\
 
    \begin{center}
     \includegraphics[width=4.5in]{img/labels2.jpg}
        \end{center}
    \caption{Comparison of controller properties in different environmental conditions. \textbf{Line plots} show the progression of the means of the population (quartiles over all runs), while \textbf{boxplots} show the distribution of the means in the final generation. Significance levels for the Wilcoxon tests in the boxplots are $* < 0.05$, $** < 0.01$, $*** < 0.001$. Note that, naturally, the values of Baseline are the same in Flat or Tilted.}
    \label{fig:controllers_gens_seasonal}
 
\end{figure}

 Because the new encoding that we propose in the current paper, i.e., \textit{Plasticoding}, allows one same individual to develop distinct morphologies and/or controllers (and thus also behavior) according the environmental conditions, we expect the system to be less impacted by  this trade-off. Therefore, here we compare two populations separately evolved in the seasonal environmental condition, a) for one population the encoding method was \textit{Baseline}, b) for another population the encoding method was  \textit{Plasticoding}. 
 Concerning morphological properties, in the Flat season robots are bigger for \textit{Plasticoding} than for \textit{Baseline}, while they also have more sensors (Fig.~\ref{fig:morphologies_gens_seasonal}). Concerning controller properties, we observe differences that directly relate to sensor differences in morphology: In the Flat season for the \textit{Plasticoding} robots have higher Sensors Reach and Recurrence (Fig.~\ref{fig:controllers_gens_seasonal}). This indicates that, in \textit{Plasticoding}, robot evolve to have  sensors sending signals to more motors, and that the brain of the robot has more memory, when compared to \textit{Baseline}. Notably, given that the neurons of the controllers are oscillators, recurrence can only make a difference if there are inputs. In simpler words, memory is only needed if there is something to remember, and this may explain why both metrics (Sensor Reach and Recurrence) are increased at the same time. Therefore, the selection pressure for higher Recurrence in the Flat environmental condition suggests that sensors are useful in the context of seasons provided that robots have capacity for \textit{phenotypic plasticity}.  

\begin{figure}[t]
 \centering
    \includegraphics[width=2in]{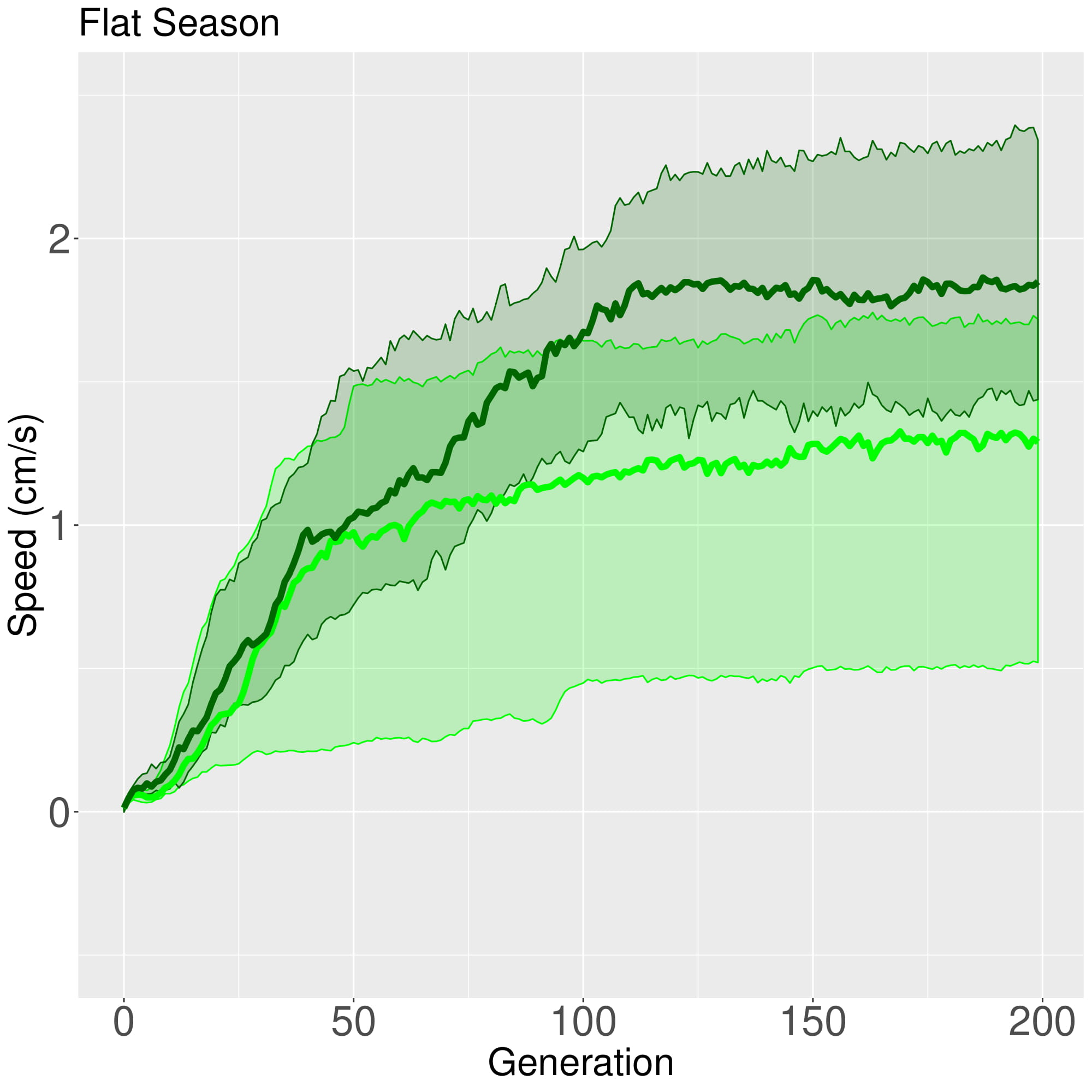}
  \includegraphics[width=2in]{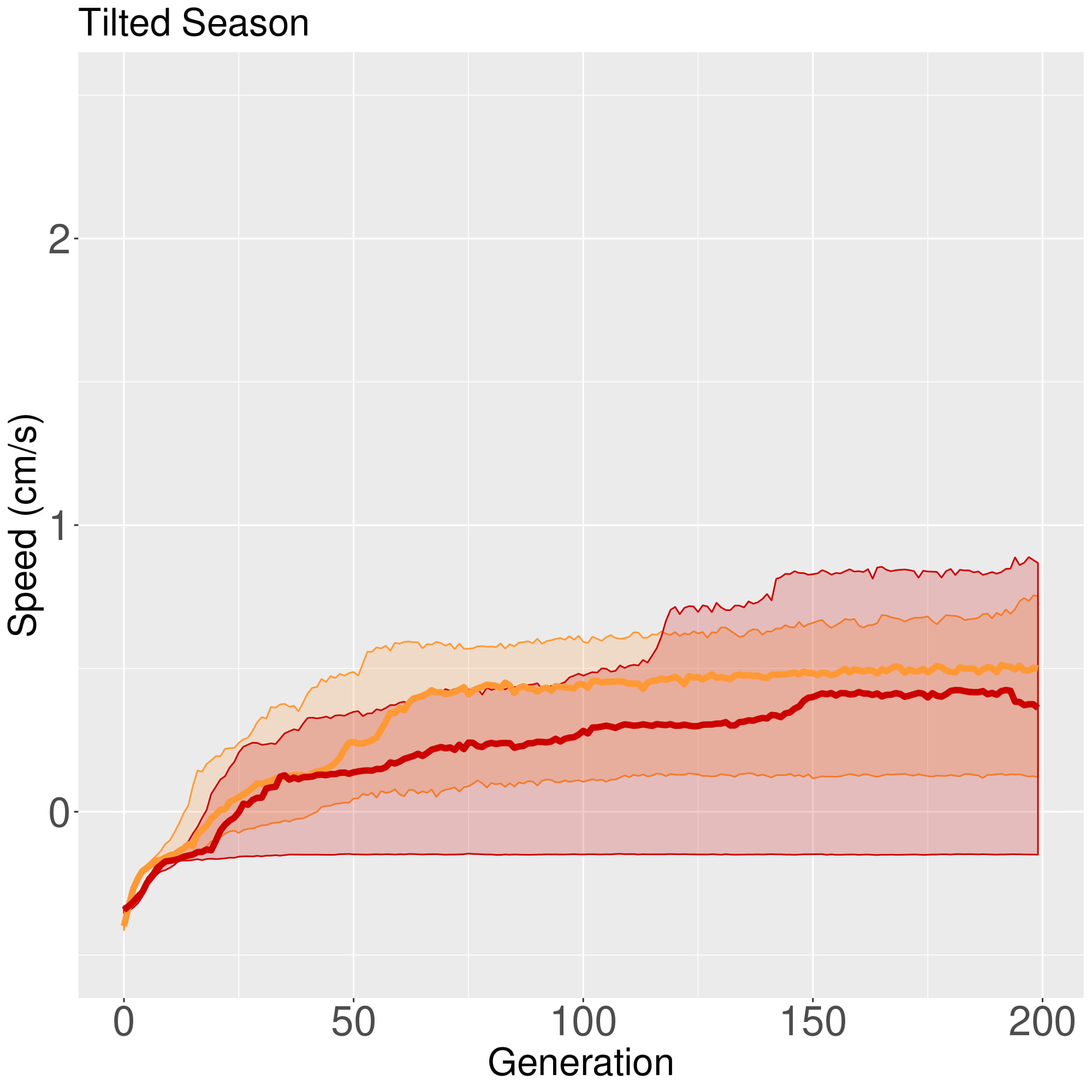}
    \includegraphics[width=1.1in]{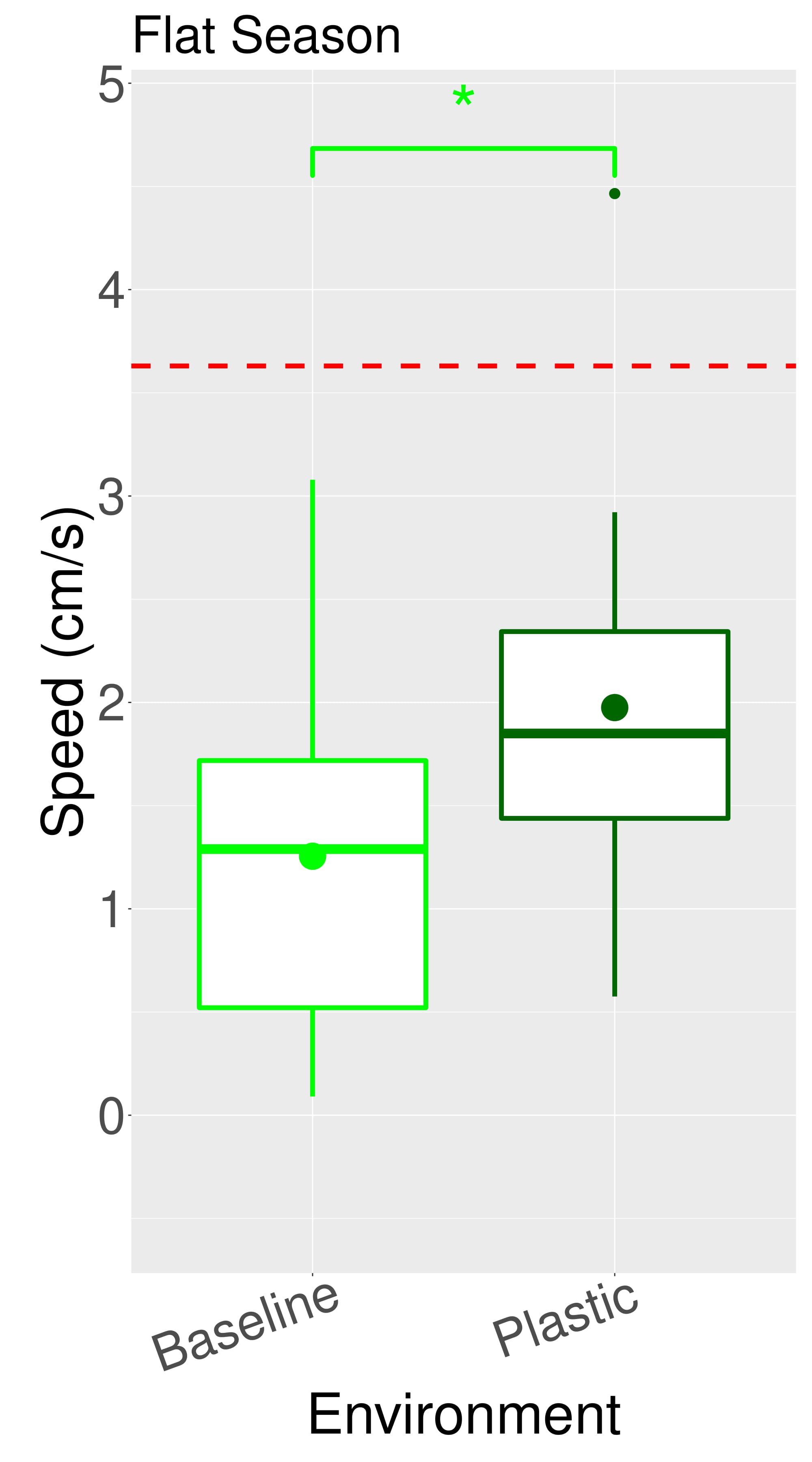}
    \includegraphics[width=1.1in]{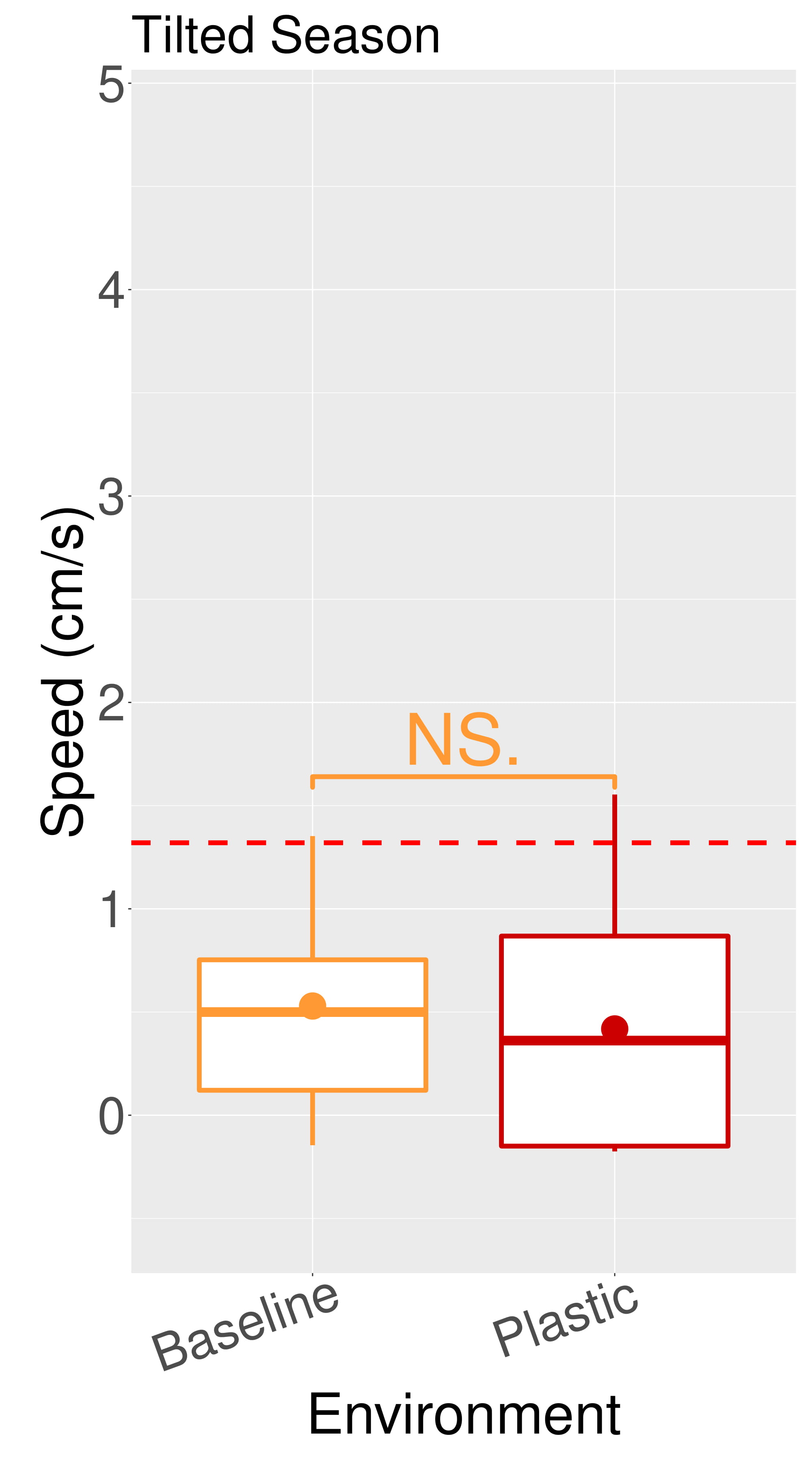}
    \\
    \includegraphics[width=2in]{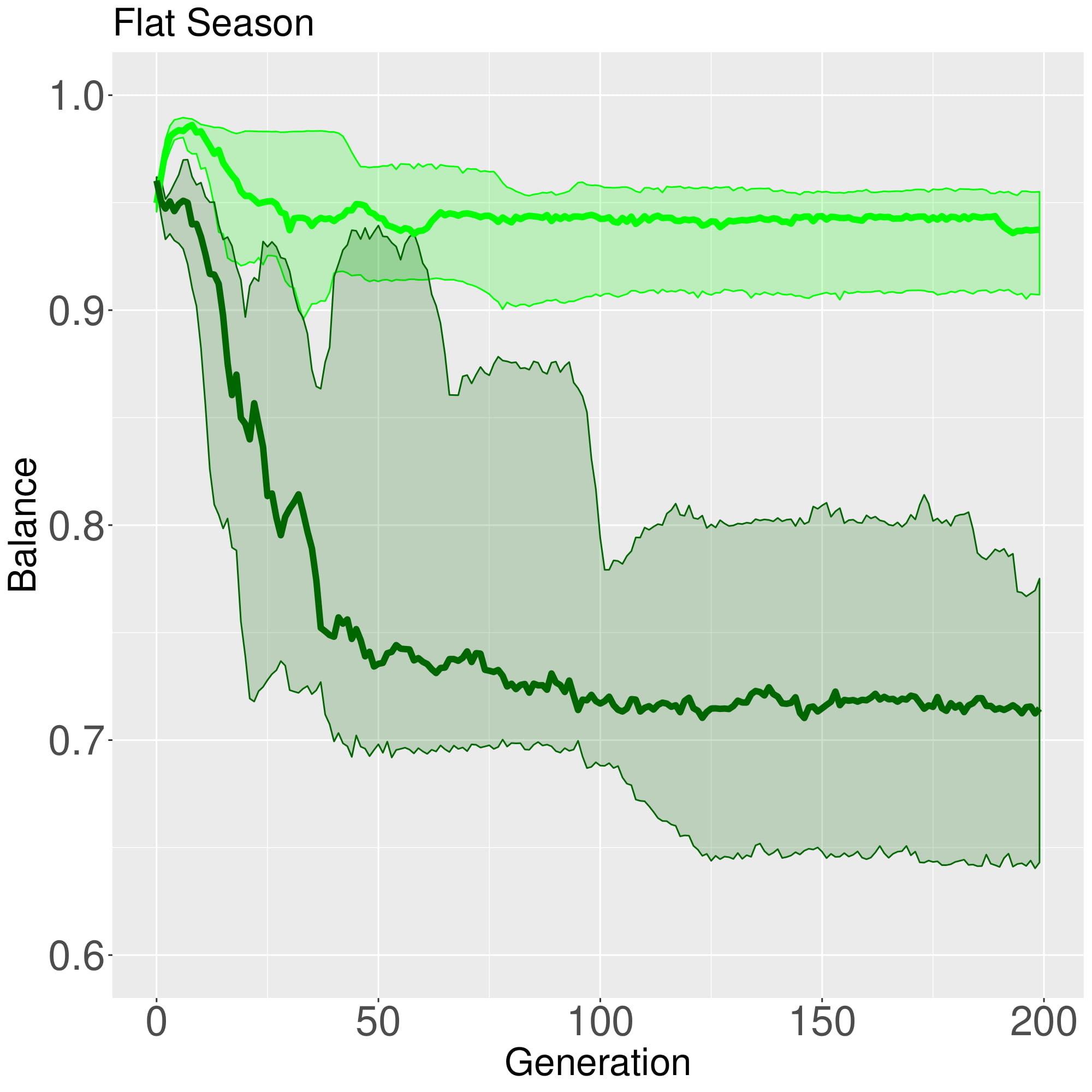}
      \includegraphics[width=2in]{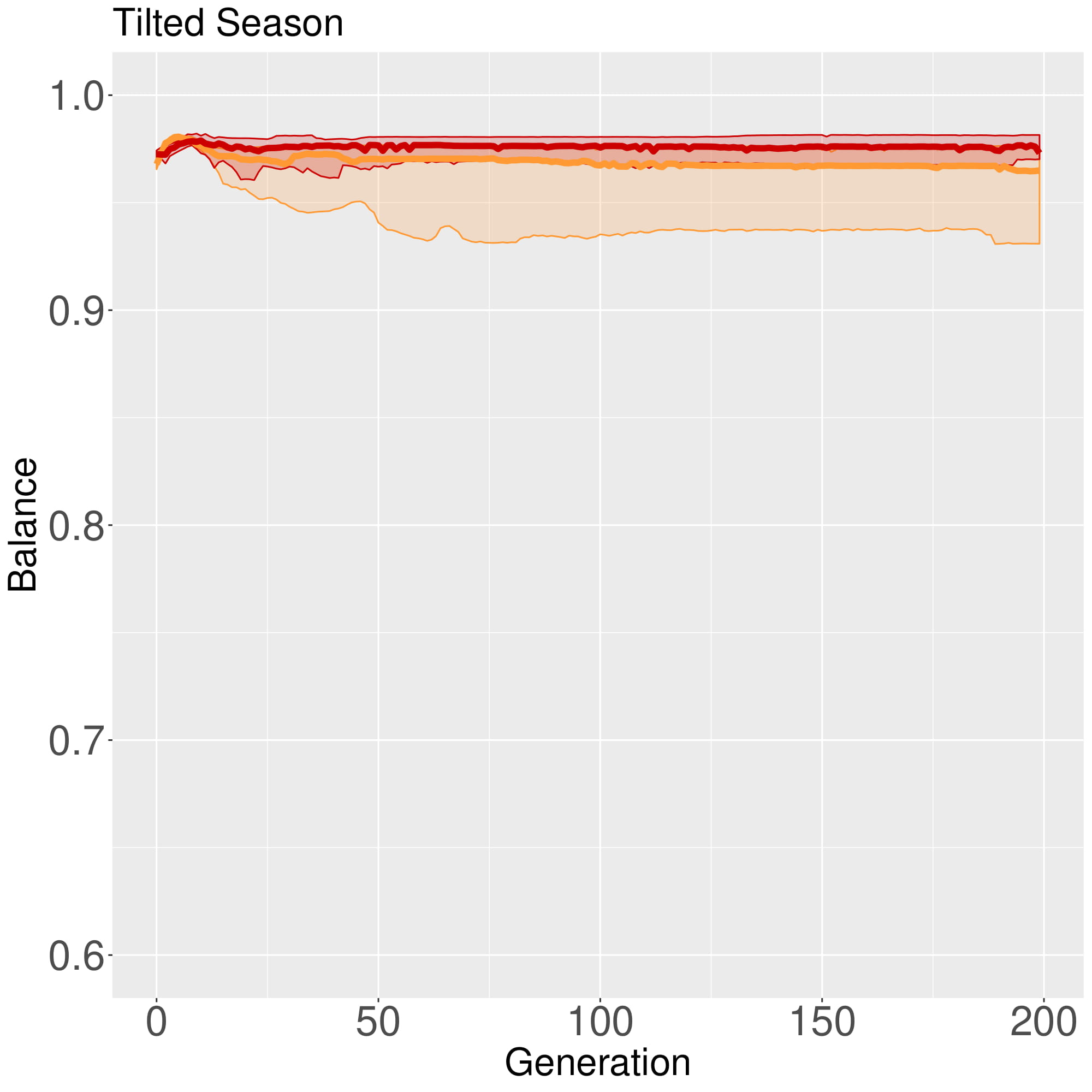}
    \includegraphics[width=1.1in]{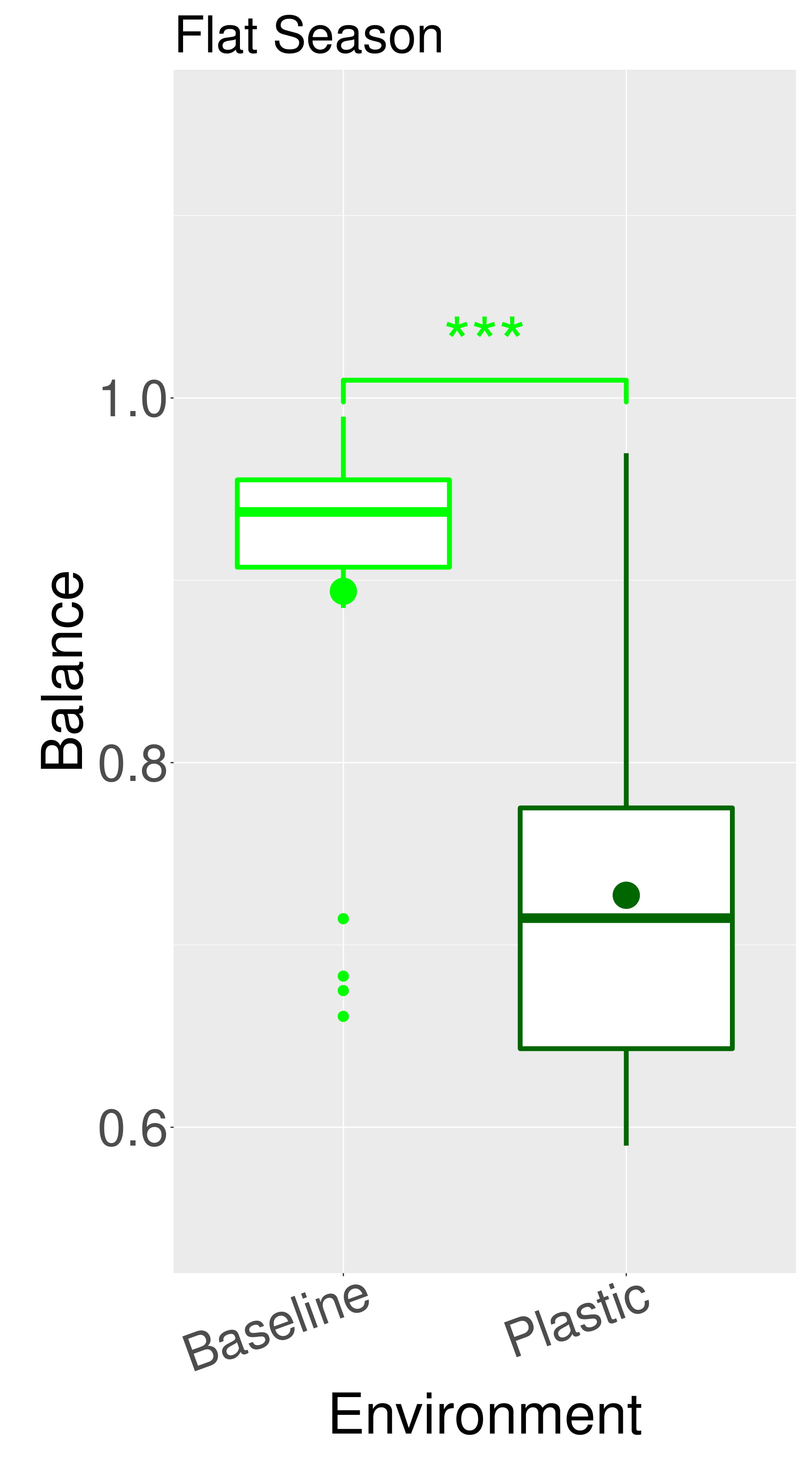}
    \includegraphics[width=1.1in]{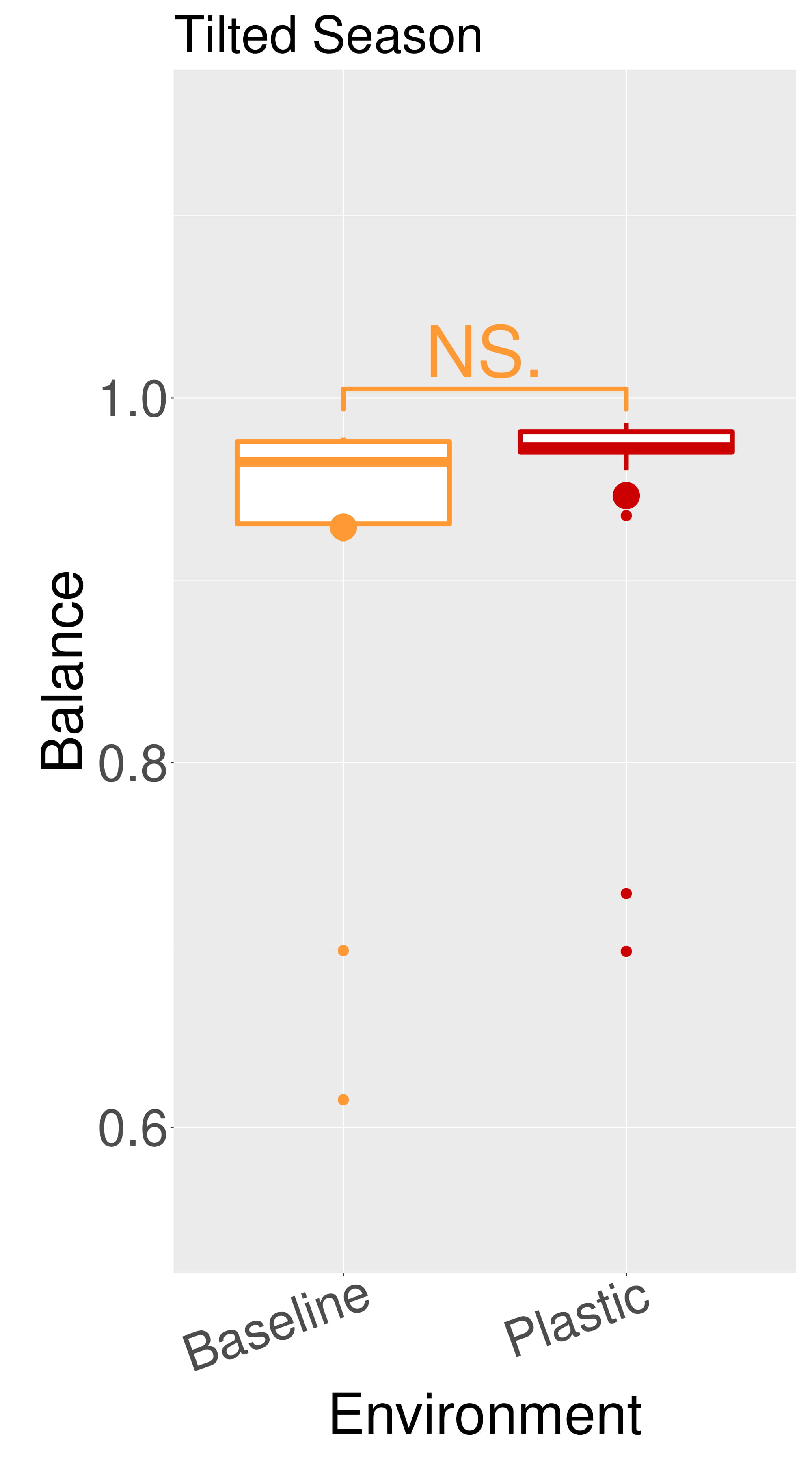}
 
 
 \includegraphics[width=4in]{img/labels2.jpg}
 
    \caption{Comparison of behavioral properties in different environmental conditions. \textbf{Line plots} show the progression of the means of the population (quartiles over all runs), while \textbf{boxplots} show the distribution of the means in the final generation. Significance levels for the Wilcoxon tests in the boxplots are $* < 0.05$, $** < 0.01$, $*** < 0.001$. The red dotted lines represent the mean Speed when evolving in an static environmental condition.}
    \label{fig:behavior_gens_seasonal}
 
\end{figure}

In the Flat environmental condition, this phenotypic differentiation is clearly reflected on the emergent behavior, i.e., behavior that emerges from the interaction among morphology, controller and environment to achieve the rewarded behavior (task). The Balance of the robots is lower for \textit{Plasticoding} than for \textit{Baseline} (Fig.~\ref{fig:behavior_gens_seasonal}), and this behavioral property agrees with their predominant gait\footnote{A video showing examples of emergent gaits is available on the link \url{https://www.youtube.com/watch?v=43wsQfWMo-Q&feature=youtu.be}}, which is rolling for \textit{Plasticoding} and rowing or dragging for \textit{Baseline}. While rolling requires the imbalancement of the center of mass of the robots, rowing and dragging require the opposite. Note that this rolling gait was expected to be observed because rolling is a common emergent behavior when evolving in a static Flat environmental condition~\citep{miras2019effects}. Nevertheless, though rolling is predominant for \textit{Plasticoding} in Flat, this is not the case for \textit{Baseline}, which delivers predominantly a gait of rowing/dragging, which are more common when evolving in a static Tilted environmental condition~\citep{miras2019effects} instead. This corroborates with our discussion in the beginning of this section, concerning a pressure for the most generalist strategy for locomotion. 

Finally, the rewarded behavior (task) shows that the phenotypic and behavioral changes caused by \textit{Plasticoding} helped to improve the performance on the task when in the Flat season. The Speed of the evolved population is $58\%$ higher for \textit{Plasticoding} than for \textit{Baseline}. This difference was proven significant with an Wilcoxon test presenting a $p$-value of $0.015$ (Fig.~\ref{fig:behavior_gens_seasonal}). It is no surprise that \textit{Baseline} delivers robots that perform worse on the task when in the Flat environmental condition, considering that the \textit{Baseline} gave in to the selection pressure existent in the Tilted environmental condition for robots that row and drag.

The red dotted lines in the boxplots of Speed (Fig.~\ref{fig:behavior_gens_seasonal}) mark a reference for a ``known achievement''. These lines represent the means of Speed when evolving populations in a static environmental condition, i.e., the environmental condition were always the same through the their lifetime, and serve as a reference of what could be achieved in a less constrained scenario. This leaves us with an open question: is it possible through \textit{phenotypic plasticity} to achieve a performance non different from when evolving in static environmental conditions, or is this degradation at least to some extent, inevitable given the costs of evolving regulatory capabilities?   

\begin{figure}[t]
 \centering
    \includegraphics[width=3.2in]{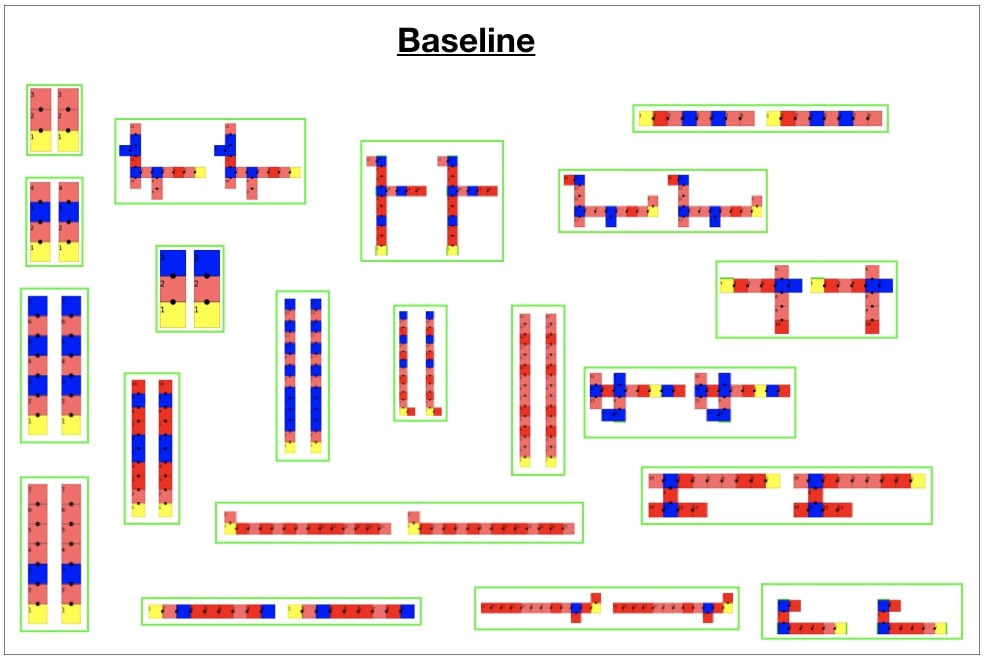}
     \includegraphics[width=3.2in]{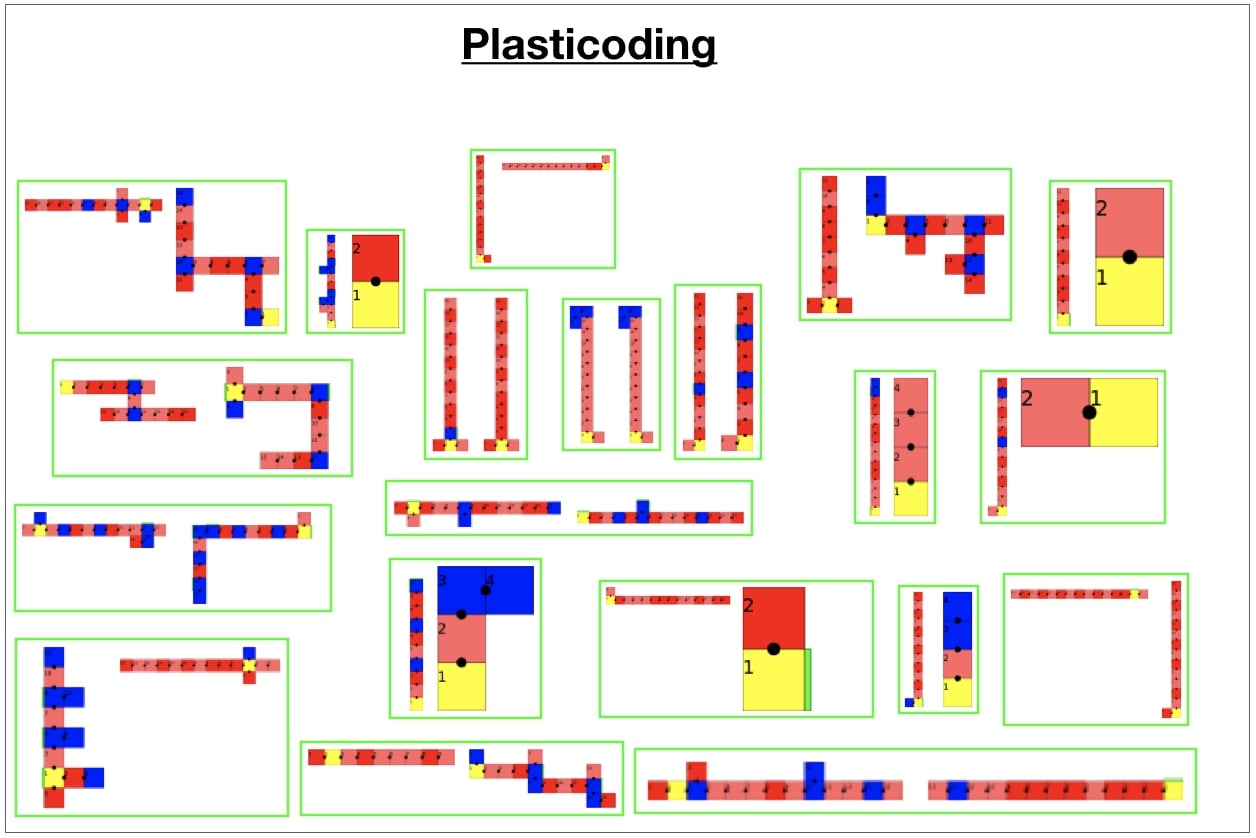}
    \caption{Best robot of each experiment repetition using each encoding method. At the \textbf{left} of each green box is the morphology developed in the \textbf{Flat season}, while at the \textbf{right} is the morphology developed in the \textbf{Tilted season}. Note that for the Baseline, naturally, in each green box both morphologies are the same. Pictures were rescaled to fit the frames accordingly.}
    \label{fig:robots_best}
 
\end{figure}

Importantly, all aforementioned differentiation between \textit{Plasticoding} and  \textit{Baseline} that took place in the Flat season did not take place in the Tilted season. This is also true for the task performance, for which no gain or loss was achieved. One possible explanation might be the fact that the Tilted environmental condition is more challenging~\citep{miras2019effects} than the Flat. In Figure~\ref{fig:morphologies_gens_seasonal}, by observing the curves of Size, we see that until around generation $25$ the search is trying to escape the local optimum mentioned in Section~\ref{sec:fitness}. That is, first the population turns into very small robots, then later on they grow bigger. However, although we see a stable increase in the average, there is a lot of variance maintained until the end on the evolutionary period. Figure~\ref{fig:robots_best} helps to illustrate that, showing that it is not uncommon to end up with very small robots, so small that they can barely locomote. Perhaps one explanation to this is that the obvious difficulty of evolving robots in the Tilted environmental condition led evolution to exploit the Flat environmental condition instead.


\section{Concluding remarks}

We investigated the effects of environmental regulation on the evolution of robots using a novel encoding method that we called \textit{Plasticoding}. This regulation gave robots a capacity for \textit{phenotypic plasticity}, so that one same robot could develop a different morphology, controller or behavior given changes in environmental conditions. In a set of experiments, we evolved robots that had to cope with two different environmental conditions during their life: one flat floor and one inclined floor. Importantly, each of these conditions presents a different selection pressure~\citep{miras2019impact}. This means that in each one of these environments, the mostly likely emergent morphological and behavioral properties are significantly different. By comparing the results achieved by \textit{Plasticoding} to a baseline encoding (similar encoding but with no regulation capacity), we showed that environmental regulation improves robot adaptation while leading to different evolved morphologies, controllers, and behavior.
 
For future work we propose to improve \textit{Plasticoding} through experimenting with a) the mutation probabilities, trying to balance changes in the production-rules versus regulation clauses; b) different methods of initialization for the production-rules and regulation clauses. Additionally, we propose to investigate effects on evolvability through: a) limiting \textit{phenotypic plasticity} to happen during morphogenesis only; b) allowing the inheritance of regulatory changes (\textit{epigenetics}).

\bibliographystyle{frontiersinSCNS_ENG_HUMS} 

\begin{thebibliography}{29}
\providecommand{\natexlab}[1]{#1}
\expandafter\ifx\csname urlstyle\endcsname\relax
  \providecommand{\doi}[1]{doi:\discretionary{}{}{}#1}\else
  \providecommand{\doi}{doi:\discretionary{}{}{}\begingroup
  \urlstyle{rm}\Url}\fi
\providecommand{\selectlanguage}[1]{\relax}
\providecommand{\bibAnnoteFile}[1]{%
  \IfFileExists{#1}{\begin{quotation}\noindent\textsc{Key:} #1\\
  \textsc{Annotation:}\ \input{#1}\end{quotation}}{}}
\providecommand{\bibAnnote}[2]{%
  \begin{quotation}\noindent\textsc{Key:} #1\\
  \textsc{Annotation:}\ #2\end{quotation}}

\bibitem[{Auerbach et~al.(2014)Auerbach, Aydin, Maesani, Kornatowski,
  Cieslewski, Heitz et~al.}]{auerbach2014robogen}
Auerbach, J., Aydin, D., Maesani, A., Kornatowski, P., Cieslewski, T., Heitz,
  G., et~al. (2014).
\newblock Robogen: Robot generation through artificial evolution.
\newblock In \emph{Artificial Life 14: Proceedings of the Fourteenth
  International Conference on the Synthesis and Simulation of Living Systems}
  (The MIT Press), 136--137
\input{auerbach2014robogen}

\bibitem[{Auerbach and Bongard(2014)}]{2014-AB}
Auerbach, J. and Bongard, J. (2014).
\newblock Environmental influence on the evolution of morphological complexity
  in machines.
\newblock \emph{PLOS Computational Biology} 10, e1003399
\input{2014-AB}

\bibitem[{Bongard(2011)}]{bongard2011morphological}
Bongard, J.~C. (2011).
\newblock Morphological and environmental scaffolding synergize when evolving
  robot controllers.
\newblock \emph{Boo Morphological and Environmental Scaffolding Synergize When
  Evolving Robot Controller}
\input{bongard2011morphological}

\bibitem[{Bossdorf et~al.(2008)Bossdorf, Richards, and
  Pigliucci}]{bossdorf2008epigenetics}
Bossdorf, O., Richards, C.~L., and Pigliucci, M. (2008).
\newblock Epigenetics for ecologists.
\newblock \emph{Ecology letters} 11, 106--115
\input{bossdorf2008epigenetics}

\bibitem[{Brawer et~al.(2017)Brawer, Hill, Livingston, Aaron, Bongard, and
  Long~Jr}]{brawer2017epigenetic}
Brawer, J., Hill, A., Livingston, K., Aaron, E., Bongard, J., and Long~Jr,
  J.~H. (2017).
\newblock Epigenetic operators and the evolution of physically embodied robots.
\newblock \emph{Frontiers in Robotics and AI} 4, 1
\input{brawer2017epigenetic}

\bibitem[{Daudelin et~al.(2018)Daudelin, Jing, Tosun, Yim, Kress-Gazit, and
  Campbell}]{daudelin2018integrated}
Daudelin, J., Jing, G., Tosun, T., Yim, M., Kress-Gazit, H., and Campbell, M.
  (2018).
\newblock An integrated system for perception-driven autonomy with modular
  robots.
\newblock \emph{Science Robotics} 3, eaat4983
\input{daudelin2018integrated}

\bibitem[{Doncieux et~al.(2015)Doncieux, Bredeche, Mouret, and
  Eiben}]{doncieux2015evolutionary}
Doncieux, S., Bredeche, N., Mouret, J.-B., and Eiben, A.~E. (2015).
\newblock Evolutionary robotics: what, why, and where to.
\newblock \emph{Frontiers in Robotics and AI} 2, 4
\input{doncieux2015evolutionary}

\bibitem[{Eiben and Smith(2015)}]{eiben2015evolutionary}
Eiben, A.~E. and Smith, J. (2015).
\newblock From evolutionary computation to the evolution of things.
\newblock \emph{Nature} 521, 476
\input{eiben2015evolutionary}

\bibitem[{Eiben et~al.(2003)Eiben, Smith et~al.}]{eiben2003introduction}
Eiben, A.~E., Smith, J.~E., et~al. (2003).
\newblock \emph{Introduction to evolutionary computing}, vol.~53 (Springer)
\input{eiben2003introduction}

\bibitem[{Fusco and Minelli(2010)}]{fusco2010phenotypic}
[Dataset] Fusco, G. and Minelli, A. (2010).
\newblock Phenotypic plasticity in development and evolution: facts and
  concepts
\input{fusco2010phenotypic}

\bibitem[{Hornby and Pollack(2001)}]{hornby2001body}
Hornby, G.~S. and Pollack, J.~B. (2001).
\newblock Body-brain co-evolution using l-systems as a generative encoding.
\newblock In \emph{Proceedings of the 3rd Annual Conference on Genetic and
  Evolutionary Computation} (Morgan Kaufmann Publishers), 868--875
\input{hornby2001body}

\bibitem[{Hupkes et~al.(2018)Hupkes, Jelisavcic, and Eiben}]{hupkes2018revolve}
Hupkes, E., Jelisavcic, M., and Eiben, A.~E. (2018).
\newblock Revolve: a versatile simulator for online robot evolution.
\newblock In \emph{International Conference on the Applications of Evolutionary
  Computation} (Springer), 687--702
\input{hupkes2018revolve}

\bibitem[{Jacob(1994)}]{jacob1994genetic}
Jacob, C. (1994).
\newblock Genetic l-system programming.
\newblock \emph{Parallel Problem Solving from Nature—PPSN III} , 333--343
\input{jacob1994genetic}

\bibitem[{Kelly et~al.(2011)Kelly, Panhuis, and Stoehr}]{kelly2011phenotypic}
Kelly, S.~A., Panhuis, T.~M., and Stoehr, A.~M. (2011).
\newblock Phenotypic plasticity: molecular mechanisms and adaptive
  significance.
\newblock \emph{Comprehensive Physiology} 2, 1417--1439
\input{kelly2011phenotypic}

\bibitem[{Kriegman et~al.(2018{\natexlab{a}})Kriegman, Cheney, and
  Bongard}]{kriegman2018morphological}
Kriegman, S., Cheney, N., and Bongard, J. (2018{\natexlab{a}}).
\newblock How morphological development can guide evolution.
\newblock \emph{Scientific reports} 8, 13934
\input{kriegman2018morphological}

\bibitem[{Kriegman et~al.(2018{\natexlab{b}})Kriegman, Cheney, Corucci, and
  Bongard}]{kriegman2018interoceptive}
Kriegman, S., Cheney, N., Corucci, F., and Bongard, J.~C. (2018{\natexlab{b}}).
\newblock Interoceptive robustness through environment-mediated morphological
  development.
\newblock \emph{arXiv preprint arXiv:1804.02257}
\input{kriegman2018interoceptive}

\bibitem[{Liknes and Swanson(2011)}]{liknes2011phenotypic}
Liknes, E.~T. and Swanson, D.~L. (2011).
\newblock Phenotypic flexibility of body composition associated with seasonal
  acclimatization in passerine birds.
\newblock \emph{Journal of Thermal Biology} 36, 363--370
\input{liknes2011phenotypic}

\bibitem[{Mills et~al.(2018)Mills, Bragina, Kumar, Zimova, Lafferty, Feltner
  et~al.}]{mills2018winter}
Mills, L.~S., Bragina, E.~V., Kumar, A.~V., Zimova, M., Lafferty, D.~J.,
  Feltner, J., et~al. (2018).
\newblock Winter color polymorphisms identify global hot spots for evolutionary
  rescue from climate change.
\newblock \emph{Science} 359, 1033--1036
\input{mills2018winter}

\bibitem[{Miras and Eiben(2019{\natexlab{a}})}]{miras2019effects}
Miras, K. and Eiben, A.~E. (2019{\natexlab{a}}).
\newblock Effects of environmental conditions on evolved robot morphologies and
  behavior.
\newblock In \emph{Proceedings of the Genetic and Evolutionary Computation
  Conference} (ACM), 125--132
\input{miras2019effects}

\bibitem[{Miras and Eiben(2019{\natexlab{b}})}]{miras2019impact}
Miras, K. and Eiben, A.~E. (2019{\natexlab{b}}).
\newblock The impact of environmental history on evolved robot properties.
\newblock In \emph{The 2018 Conference on Artificial Life: A Hybrid of the
  European Conference on Artificial Life (ECAL) and the International
  Conference on the Synthesis and Simulation of Living Systems (ALIFE)} (MIT
  Press), 396--403
\input{miras2019impact}

\bibitem[{Miras et~al.(2018{\natexlab{a}})Miras, Haasdijk, Glette, and
  Eiben}]{2miras2018}
Miras, K., Haasdijk, E., Glette, K., and Eiben, A.~E. (2018{\natexlab{a}}).
\newblock {Effects of Selection Preferences on Evolved Robot Morphologies and
  Behaviors}.
\newblock In \emph{Proceedings of the Artificial Life Conference 2018 (ALIFE
  2018)}, eds. T.~Ikegami, N.~Virgo, O.~Witkowski, R.~Suzuki, M.~Oka, and
  H.~Iizuka (Tokyo: MIT Press), 224--231
\input{2miras2018}

\bibitem[{Miras et~al.(2018{\natexlab{b}})Miras, Haasdijk, Glette, and
  Eiben}]{mirassearch2017}
Miras, K., Haasdijk, E., Glette, K., and Eiben, A.~E. (2018{\natexlab{b}}).
\newblock Search space analysis of evolvable robot morphologies.
\newblock In \emph{Applications of Evolutionary Computation - 21st
  International Conference, EvoApplications 2018} (Springer), vol. 10784 of
  \emph{Lecture Notes in Computer Science}. 703--718
\input{mirassearch2017}

\bibitem[{Nolfi et~al.(2016)Nolfi, Bongard, Husbands, and
  Floreano}]{nolfi2016evolutionary}
Nolfi, S., Bongard, J., Husbands, P., and Floreano, D. (2016).
\newblock Evolutionary robotics.
\newblock In \emph{Springer Handbook of Robotics} (Springer). 2035--2068
\input{nolfi2016evolutionary}

\bibitem[{Nolfi and Floreano(2000)}]{nolfi2000evolutionary}
Nolfi, S. and Floreano, D. (2000).
\newblock \emph{Evolutionary Robotics: The Biology, Intelligence, and
  Technology of Self-Organizing Machines}
\input{nolfi2000evolutionary}

\bibitem[{Pfeifer and Iida(2005)}]{pfeifer2005morphological}
Pfeifer, R. and Iida, F. (2005).
\newblock Morphological computation: Connecting body, brain, and environment.
\newblock In \emph{Creating Brain-Like Intelligence} (Springer), vol. 5436.
  130--136
\input{pfeifer2005morphological}

\bibitem[{Price et~al.(2003)Price, Qvarnstr{\"o}m, and Irwin}]{price2003role}
Price, T.~D., Qvarnstr{\"o}m, A., and Irwin, D.~E. (2003).
\newblock The role of phenotypic plasticity in driving genetic evolution.
\newblock \emph{Proceedings of the Royal Society of London. Series B:
  Biological Sciences} 270, 1433--1440
\input{price2003role}

\bibitem[{Rothlauf(2006)}]{rothlauf2006representations}
Rothlauf, F. (2006).
\newblock Representations for genetic and evolutionary algorithms.
\newblock In \emph{Representations for Genetic and Evolutionary Algorithms}
  (Springer). 9--32
\input{rothlauf2006representations}

\bibitem[{Sapolsky(2017)}]{sapolsky2017behave}
Sapolsky, R.~M. (2017).
\newblock \emph{Behave: The biology of humans at our best and worst} (Penguin)
\input{sapolsky2017behave}

\bibitem[{Sims(1994)}]{sims1994evolving}
Sims, K. (1994).
\newblock Evolving 3d morphology and behavior by competition.
\newblock \emph{Artificial life} 1, 353--372
\input{sims1994evolving}

\end{thebibliography}



\end{document}